\documentclass[10pt,twocolumn,letterpaper]{article}

\usepackage{cvpr}
\usepackage{times}
\usepackage{epsfig}
\usepackage{graphicx}
\usepackage{amsmath}
\usepackage{amssymb}
\usepackage[caption=false,font=footnotesize]{subfig}

\usepackage[breaklinks=true,bookmarks=false]{hyperref}

\cvprfinalcopy 

\def\cvprPaperID{1352} 

\ifcvprfinal\fi
\begin{document}

\title{Pedestrian Detection Inspired by Appearance Constancy and Shape Symmetry}

\author{Jiale Cao\textsuperscript{a}, Yanwei Pang\textsuperscript{a*}, and Xuelong Li\textsuperscript{b}\\
\textsuperscript{a}~School of Electronic Information Engineering, Tianjin University, Tianjin 300072, China
\\
\textsuperscript{b}~Institute of Optics and Precision Mechanics, Chinese Academy of Sciences, Xi'an 710119, China\\
{\tt\small \{connor,pyw\}@tju.edu.cn \qquad xuelong{\_}li@ieee.org}
}

\maketitle

\begin{abstract}
The discrimination and simplicity of features are very important for 
effective and efficient pedestrian detection. However, most state-of-the-art 
methods are unable to achieve good tradeoff between accuracy and efficiency. 
Inspired by some 
simple inherent attributes of pedestrians (i.e., appearance constancy and 
shape symmetry), we propose two new types of non-neighboring features (NNF): 
side-inner difference features (SIDF) and symmetrical similarity features 
(SSF). SIDF can characterize the difference between the background and 
pedestrian and the difference between the pedestrian contour and its inner 
part. SSF can capture the symmetrical similarity of pedestrian shape. 
However, it's difficult for neighboring features to have such above 
characterization abilities. Finally, we propose to combine both 
non-neighboring and neighboring features for pedestrian detection. It's 
found that non-neighboring features can further decrease the average miss 
rate by 4.44{\%}. Experimental results on INRIA and Caltech pedestrian 
datasets demonstrate the effectiveness and efficiency of the proposed 
method. Compared to the state-of-the-art methods without using CNN, our 
method achieves the best detection performance on Caltech, outperforming the 
second best method (i.e., Checkboards) by 1.63{\%}.
\end{abstract}

\section{Introduction}
Pedestrian detection is a premise in many computer vision tasks 
including gait recognition, behavior analysis, action recognition, and 
camera-based driver assistance. Generally speaking, the performance of 
pedestrian detection is determined by the performance of feature extraction 
and classification. This paper focuses on feature extraction. 

There are three manners for feature extraction: (1) completely Hand-Crafted 
(HC) features, (2) Hand-Crafted candidate features followed by Learning 
Algorithms (HCLA), and (3) Deep Leaning 
(DL) based features. Due to simplicity and robustness, it is much more 
possible for HCLA to achieve good tradeoff between efficiency and accuracy. So this paper concentrates 
on HCLA. 

Usually, the input of HCLA for pedestrian detection is CIE-LUV color channels, 
gradient histogram channels, gradient magnitude channel, and so on. Once the 
channels are specified, the question remained is how to generate candidate features 
from the channels. Most of the state-of-the-art methods generate the 
candidate features by using local (e.g., local mean features) or neighboring 
features (e.g., haar features). In fact, some inherent attributes of 
pedestrians can also be used for feature design. Inspired by appearance 
constancy and shape symmetry of pedestrians, we design two types of 
non-neighboring features for pedestrian detection: side-inner difference 
features (SIDF) and symmetrical similarity features (SSF). The contributions 
of the paper are as follows:

1) Appearance constancy and shape symmetry can be seen as the inherent attributes of 
pedestrians. Inspired by these attributes, we propose side-inner difference 
features (SIDF) and symmetrical similarity features (SSF), respectively. 
Compared to some state-of-the-art features, our features are oriented 
non-neighboring features. SIDF can characterize the difference between the background 
and pedestrian and the difference between the pedestrian contour and its 
inner part. SSF can capture the symmetrical similarity of pedestrian shape. 
However, it's difficult for neighboring features to have such above 
characterization abilities. 

2) We propose to employ non-neighboring and neighboring features for 
pedestrian detection. Among all the selected features, about 70{\%}  are neighboring features and 30{\%} of them are 
non-neighboring ones. So the non-neighboring features are complementary to 
the neighboring ones.

3) Compared to the state-of-the-art methods without using CNN, we achieve 
the best detection performance (i.e., 16.84{\%} miss rate on Caltech). 
Meanwhile, our methods achieve the best performance tradeoff between 
detection efficiency and log-average miss rate only by common CPU. Moreover, 
SIDF and SSF may also be combined with CNN features to further boost the 
detection performance.

The rest of the paper is organized as follows. We review related work in 
Section 2. The proposed method is given in Section 3. Experimental 
results are provided in Section 4. We then conclude in Section 5.

\section{Related work}
Pedestrian detection methods can be divided into three families  \cite{Benenson_TenYears_ECCV_2014}: 
DPM (Deformable Part Detectors) variants \cite{Felzenszwalb_CascadeDPM_CVPR_2010,Felzenszwalb_DPM_CVPR_2008,Ouyang_SingAid_CVPR_2013},  deep networks \cite{Girshick_RCNN_CVPR_2014,Hosang_DeepLook_CVPR_2015,Sermanet_PedUMFL_CVPR_2013} and decision 
forests \cite{Appel_Pruning_ICML_2013,Dollar_ICF_BMVC_2009,Viola_RoFace_IJCV_2004}. Our method can be categorized into the family of decision forests. Specifically, the 
process of this kind of methods is as follows: 1) a set of channel images 
are generated from an input image; 2) then, features are extracted from 
patches of the channels; and 3) finally, the features are fed into a 
decision forest learned via AdaBoost \cite{Zhang_FCF_CVPR_2015}. Feature extraction is a very important step.

Integral Channel Features (ICF) \cite{Dollar_ICF_BMVC_2009} is one of the most successful feature 
extraction method four years after Histograms of Oriented Gradients 
(HOG) \cite{Dalal_HOG_CVPR_2005} was proposed. In ICF, Dollar \textit{et al.} \cite{Dollar_ICF_BMVC_2009} proposed to combine 
three types of channels: LUV color channels, normalized gradient magnitude, and histogram of 
oriented gradients (6 channels). First-order and 
higher-order features are then generated from the channel images \cite{Dollar_ICF_BMVC_2009} . Soft 
cascade \cite{Bourdev_SoftCascade_CVPR_2005,Zhang_MIP_NIPS_2008} is then used for learning 
discriminative features \cite{Dollar_ICF_BMVC_2009}. Note that ICF is also known as ChnFtrs. 

Aggregated Channel Features (ACF) \cite{Dollar_ACF_PAMI_2014}, SquaresChnFtrs \cite{Benenson_SquareChns_CVPR_2013}, InformedHaar \cite{Zhang_Info.Haar_CVPR_2014}, Locally Decorrelated 
Channel Features (LDCF) \cite{Nam_LDCF_NIPS_2014}, and Checkboards \cite{Zhang_FCF_CVPR_2015} employ the same channel images as ICF. In ACF, 
the pixel sum of each block in each channel is computed and then 
the resulting lower resolution channels are smoothed \cite{Dollar_ACF_PAMI_2014,Dollar_FastestWest_BMVC_2010}. SquaresChnFtrs \cite{Benenson_SquareChns_CVPR_2013} is simpler 
than ICF because only the local sum of squares in each channel image is used as 
features. InformedHaar \cite{Zhang_Info.Haar_CVPR_2014} is specifically 
designed for pedestrian detection where a pool of rectangular templates is 
tailored to the statistical model of the up-right human body across the 
channels. By using the technique of Linear Discriminant Analysis (LDA) \cite{Hariharan_DDCC_ECCV_2012}, the LDCF 
features are decorrelated so that they are suited for orthogonal decision 
trees \cite{Nam_LDCF_NIPS_2014}. The decorrelation can be achieved by convolution with a filter 
bank learned by LDA. Checkboards \cite{Zhang_FCF_CVPR_2015} generalizes ICF by using filter banks to compute 
features from channel images. Six types of filters are 
considered: InformedFilters, CheckerboardsFilters, RandomFilters, 
SquaresChntrs filters, LDCF8 filters, and PcaForeground filters. 

SpatialPooling+ \cite{Paisitkriangkrai_SpatialPool_arXiv_2014,Paisitkriangkrai_SpatialPool_ECCV_2014} does not take channel images as input. Instead, it applies 
the operator of spatial pooling (e.g., max-pooling) on covariance descriptor 
and Local Binary Pattern (LBP). 

According to \cite{Benenson_TenYears_ECCV_2014} and our experimental results, the performance of the above 
methods can be summarized as follows: On the Caltech pedestrian dataset 
\cite{Caltech,Dollar_PD_PAMI_2012}, the miss rates of the above 
methods are ICF $>$ ACF $>$ SquaresChnFtrs $>$ InformedHaar $>$ LDCF $>$ SpatialPooling+ $>$ 
Checkboards. Loosely speaking, the detection speeds of these 
methods are SpatialPooling+ $<$ ICF $<$ SquaresChnFtrs $<$ Checkboards $<$ InformedHaar $<$ LDCF $<$ 
ACF. It can be concluded that no method can simultaneously obtain the best 
log-average miss rate and detection speed. That is, these methods are unable 
to achieve satisfying tradeoff between accuracy and efficiency. 

Recently, the methods based on CNN have achieved very good performance \cite{Cai_DeepPed_ICCV_2015,Luo_SDN_CVPR_2014,Tian_DeepParts_ICCV_2015,Tian_Ta_CVPR_2015,Yang_CCF_ICCV_2015}. For example, Tian \textit{et al.} \cite{Tian_DeepParts_ICCV_2015} proposed DeepParts to improve the
detection performance by handling occlusion
with an extensive part pool. Though the methods based CNN can achieve 
the best performance, it needs the relatively expensive device (i.e., GPU). 
On the other hand, the simple feature design can also be complementary to 
CNN. For example, by combining the simple local features (e.g, ACF \cite{Dollar_ACF_PAMI_2014}, Checkboards \cite{Zhang_FCF_CVPR_2015}, and LDCF \cite{Nam_LDCF_NIPS_2014}) and 
very deep CNN features (e.g., VGG \cite{Simonyan_VGG_arxiv_2014} and AlexNet \cite{Krizhevsky_AlexNet_NIPS_2012}) , Cai \textit{et al.} \cite{Cai_DeepPed_ICCV_2015} could decrease the miss 
rate from 18.9{\%} to 11.70{\%}. So in this paper, we focus the feature design in the traditional methods.

\section{Our methods}
\subsection{Appearance constancy and shape symmetry}
\begin{figure}[!t]
\centering
\subfloat[]{\includegraphics[]{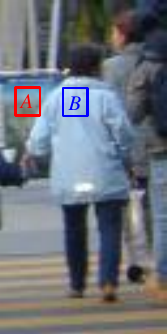}
\label{FigSomPed(a)}}
\hfil
\subfloat[]{\includegraphics[]{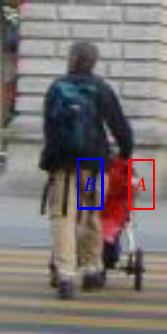}
\label{FigSomPed(b)}}
\hfil
\subfloat[]{\includegraphics[]{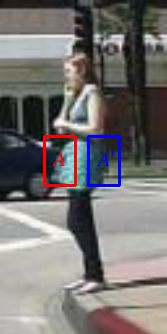}
\label{FigSomPed(c)}}
\hfil
\subfloat[]{\includegraphics[]{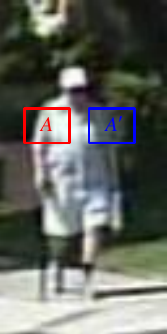}
\label{FigSomPed(d)}}
\caption{Some examples of the cropped pedestrians.}
\label{FigSomPed}
\end{figure}

Most state-of-the-art features for pedestrian detection are designed to 
describe the local image region. Thus, they don't make full use of the 
inherent attributes of pedestrians. In fact, some inherent attributes of 
pedestrians can be used to further improve detection performance. For 
example, Zhang \textit{et al.} \cite{Zhang_Info.Haar_CVPR_2014} incorporate the common sense that pedestrians usually 
appear up-right into the design of simple and efficient haar-like features. 
In this paper, we incorporate the appearance constancy and shape symmetry 
into the feature design. First of all, we give the explanations of 
appearance constancy and shape symmetry. Fig. \ref{FigSomPed} gives some examples of the 
cropped pedestrians.

1) \textit{Appearance Constancy.} The appearances of pedestrians are usually contrast to the surrounding 
background. Meanwhile, pedestrians can be seen as three main different parts 
(i.e., head, upper body, and legs). The appearances inside these parts are 
usually constancy. For example, the woman wears the sky-blue coat and black 
pants in Fig. \ref{FigSomPed}(a). We call this inherent attribute of pedestrians 
\textit{appearance constancy}. 
Thus, the regions located inside the pedestrians (e.g., patches $B$ in Figs. \ref{FigSomPed}(a) and (b)) are contrast to that located in the background (e.g., patches $A$ in Figs. \ref{FigSomPed}(a) and (b). Note that patches $A$ and $B$ lie in the same horizontal. Patches $B$ are called the inner patches, and patches $A$ are called the side patches. 

2) \textit{Shape Symmetry.} As stated in \cite{Zhang_Info.Haar_CVPR_2014}, pedestrians usually appear up-right. Thus, the pedestrian shape is loosely symmetrical in the horizontal direction. For 
example, the symmetrical region (patches $A$ and ${A}')$ in the Figs. \ref{FigSomPed}(c) 
and (d) have the similar characteristic. This inherent attribute is called 
\textit{shape symmetry}.

\begin{figure}[!t]
\centering
\subfloat[]{\includegraphics[]{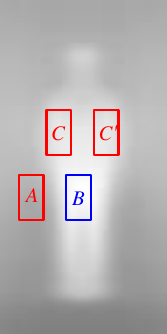}
\label{FigAvgPed(a)}}
\hfil
\subfloat[]{\includegraphics[]{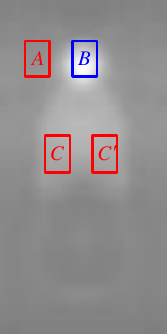}
\label{FigAvgPed(b)}}
\hfil
\subfloat[]{\includegraphics[]{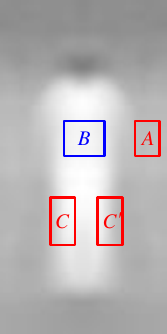}
\label{FigAvgPed(c)}}
\hfil
\subfloat[]{\includegraphics[]{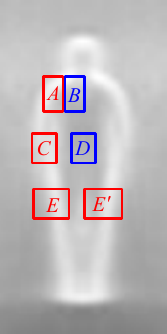}
\label{FigAvgPed(d)}}
\caption{Average values of channel images . (a) Inversed L (Luminance) channel. (b) U channel. (c) Inversed V channel. (d) G channel.}
\label{FigAvgPed}
\end{figure}

Inspired by the above appearance constancy and shape symmetry, we can design 
two types of non-neighboring features. It can be explained by the average 
appearance of pedestrians in channel images such as L, U, V, and G. Fig. \ref{FigAvgPed} 
gives the average values of above four channel images. Due to the appearance constancy, the pixel values of 
pedestrians in L, U, and V channel images are similar in the same horizontal, which are different from that of the two-side regions. Meanwhile, 
the pixel values of the inner part of pedestrians in G channel image are 
constantly small, and the pixel values of pedestrian contour in G channel 
image are relatively large. Thus, the large difference in G channel image 
can be characterized by not only the neighboring feature formed by patches 
$A$ and $B$ but also the non-neighboring feature formed by patches $C$ and 
$D$ in Fig. \ref{FigAvgPed}(d). Though there is little difference between the inner part and contour in V 
channel (Fig. \ref{FigAvgPed}(c)), the difference between the inner part and its two side 
background can be 
characterized by the non-neighboring feature formed by patches 
$A$ and $B$. Due to shape symmetry, the symmetrical regions of pedestrians in the horizontal direction have the similar characteristic. For example, the 
symmetrical patches $E$ and $E'$ in Fig. \ref{FigAvgPed}(d) 
describe the similar edge characteristic, while patches 
$C$ and $C'$ in Fig. \ref{FigAvgPed}(c) are both bright. Figs. \ref{FigAvgPed}(a) and (b) also support the above two conclusions.

\begin{figure}[!t]
\centering
\includegraphics[]{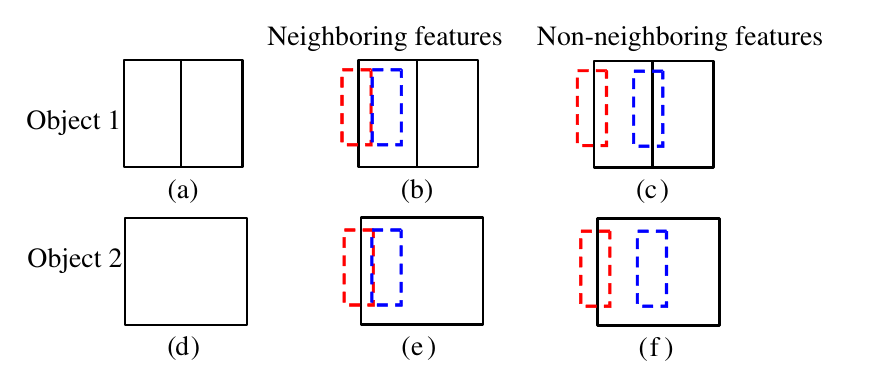}
\caption{Demonstration of the dismicrination and usefulness of non-neighboring features. (a) and (d) show Object 1 and Object 2, respectively. In (b) and (e), neighboring features are extracted. In (c) and (f), non-neighboring features are extracted.}
\label{FigDimicrinationNNF}
\end{figure}

The discrimination and usefulness of non-neighboring features are 
graphically supported by Fig. \ref{FigDimicrinationNNF}. In Fig. \ref{FigDimicrinationNNF}, there are two objects (classes) 
to be classified. We call the object in Fig. \ref{FigDimicrinationNNF}(a) Object 1 and the object in 
Fig. \ref{FigDimicrinationNNF}(d) Object 2. There is a line 
in the middle of Object 1 whereas the inner part of Object 2 is flat. In both Figs. \ref{FigDimicrinationNNF}(b) and (e), two neighboring 
dashed rectangles form a feature. We can see that this neighboring feature 
is unable to distinguish between Object 1 and Object 2 because the values of 
neighboring features in Object 1 (i.e., Fig. \ref{FigDimicrinationNNF}(b)) and Object 2 (i.e., Fig. 
\ref{FigDimicrinationNNF}(e)) are equal. Now we use in Figs. \ref{FigDimicrinationNNF}(c) and (f) two non-neighboring 
patches to form a feature. Because the blue dashed patch in Fig. \ref{FigDimicrinationNNF}(c) 
contains a line whereas the blue dashed patch in Fig. \ref{FigDimicrinationNNF}(f) contains 
nothing, the non-neighboring features in Object 1 (i.e., Fig. \ref{FigDimicrinationNNF}(c)) and 
Object 2 (i.e., Fig. \ref{FigDimicrinationNNF}(f)) have different values. The two objects can be 
correctly classified according to the different values. This demonstrates 
the dismicrination and usefulness of non-neighboring features. 

\subsection{Side-inner difference features inspired by appearance constancy}
\begin{figure}[!t]
\centering
\subfloat[Varying distance between two patches.]{\includegraphics[height=0.55in]{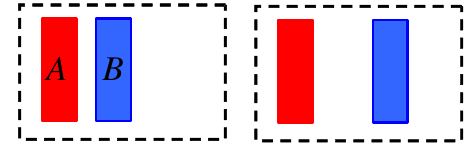}
\label{FigFormsSIDF(a)}}
\hfil
\subfloat[Varying size of two patches.]{\includegraphics[height=0.55in]{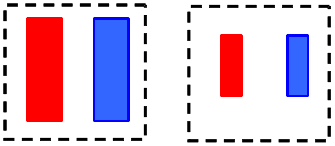}
\label{FigFormsSIDF(b)}}
\hfil
\subfloat[Varying size of one patch with the other fixed.]{\includegraphics[height=0.55in]{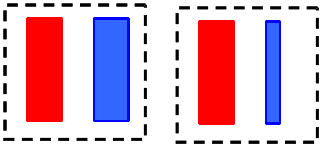}
\label{FigFormsSIDF(c)}}
\hfil
\subfloat[Varying aspect ratio.]{\includegraphics[height=0.55in]{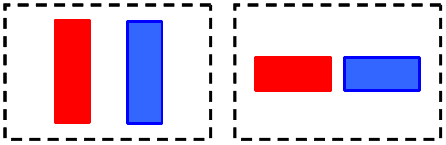}
\label{FigFormsSIDF(d)}}
\caption{Some possible forms of side-inner difference features (SIDF). }
\label{FigFormsSIDF}
\end{figure}

\begin{figure}[!t]
\centering
\subfloat[]{\includegraphics[width=0.7in]{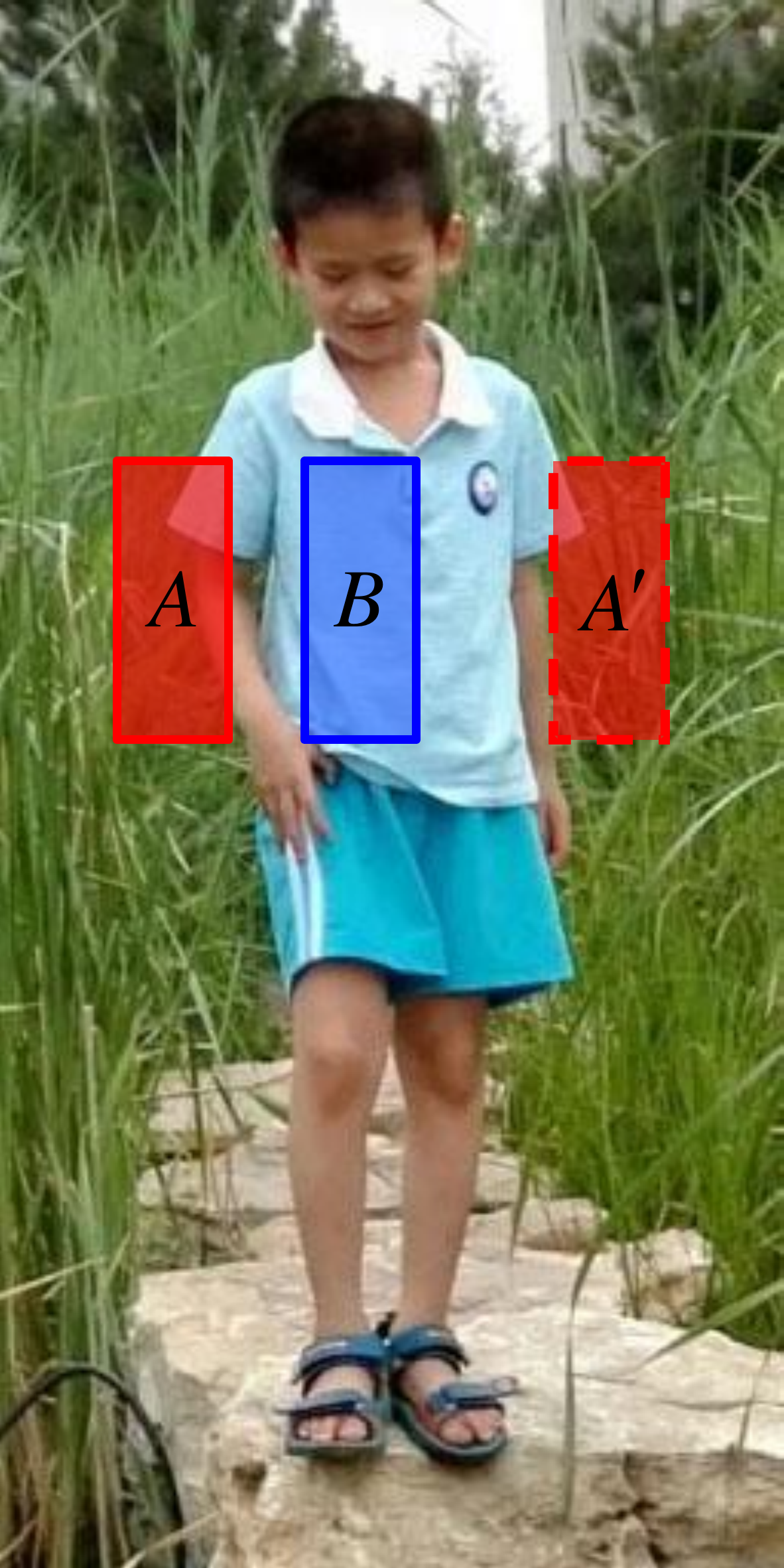}
\label{FigLocBSIDF(a)}}
\hfil
\subfloat[]{\includegraphics[width=0.7in]{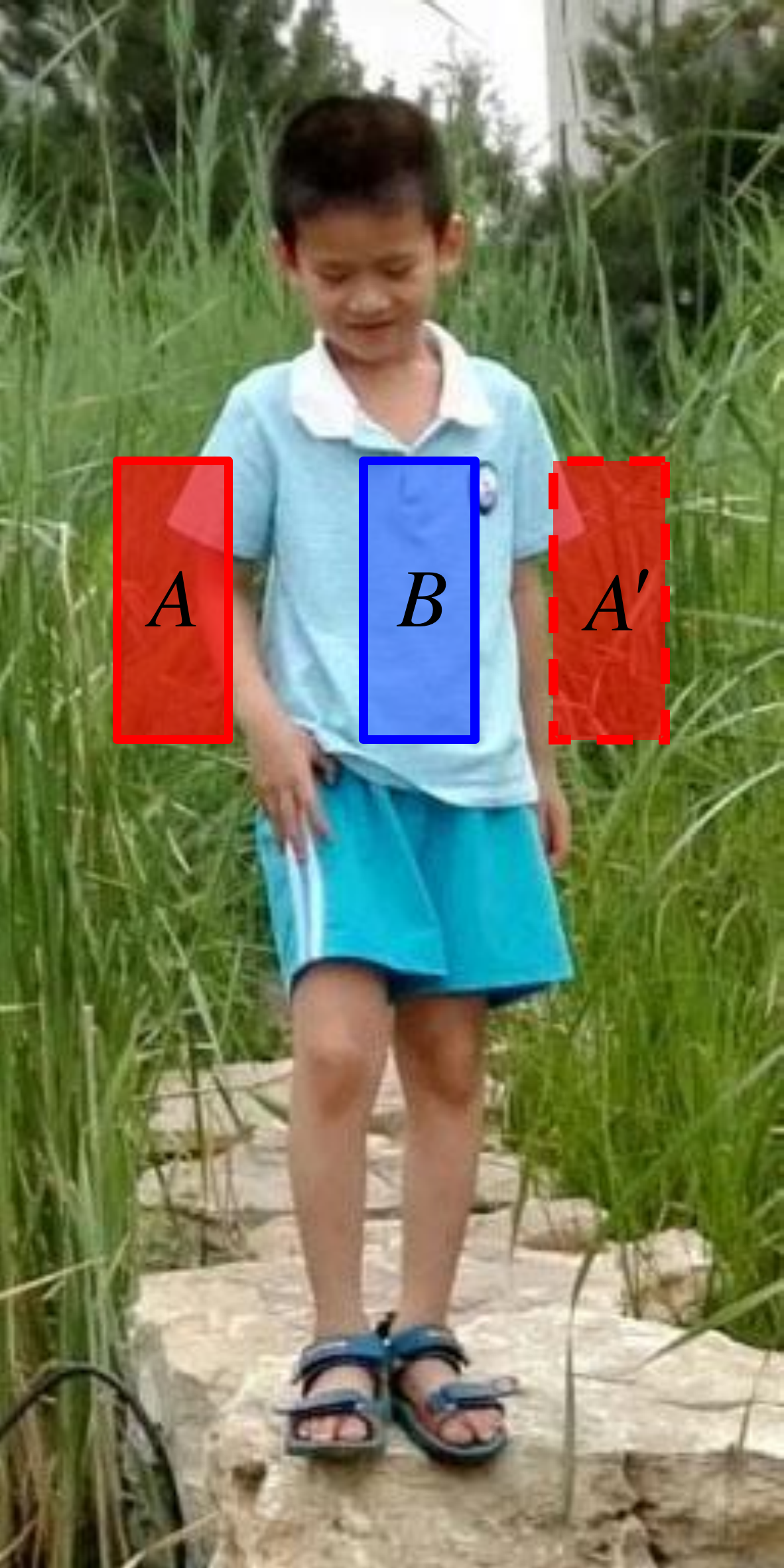}
\label{FigLocBSIDF(b)}}
\hfil
\subfloat[]{\includegraphics[width=0.7in]{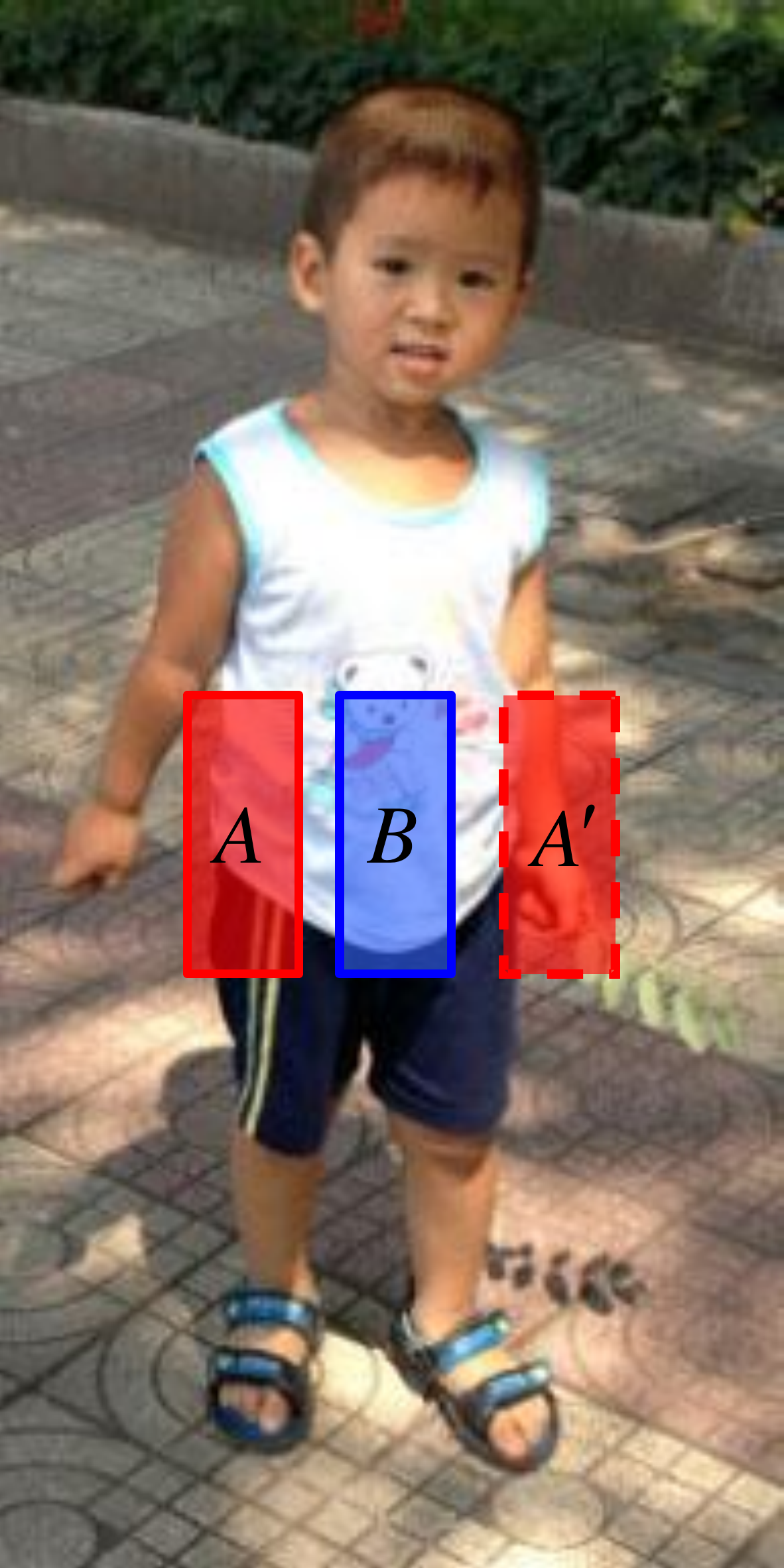}
\label{FigLocBSIDF(c)}}
\hfil
\subfloat[]{\includegraphics[width=0.7in]{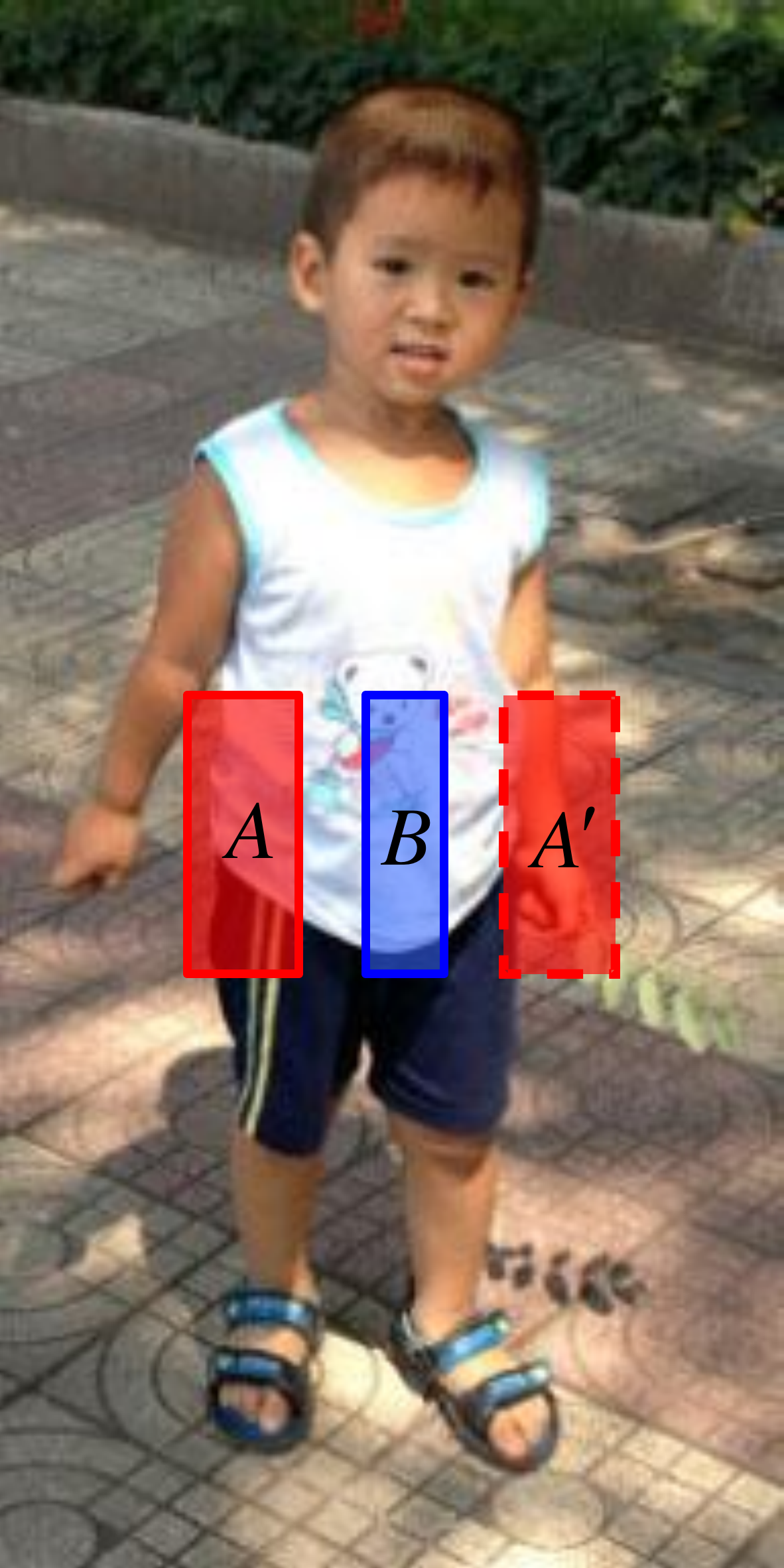}
\label{FigLocBSIDF(d)}}
\caption{The patch $B$ is randomly located between the patch $A$ and its horizontal mirror $A'$. The locations of patch $B$ in (a) and (b) are different. But they are both among $A$ and its mirror $A'$. (c) and (d) show that the width of patch $B$ can be changed.  }
\label{FigLocBSIDF}
\end{figure}

Inspired by appearance constancy, we design the non-neighboring difference 
features in the same horizontal. We call this oriented non-neighboring 
difference features Side-Inner Difference Features (SIDF). Fig. \ref{FigFormsSIDF} gives some 
possible forms of SIDF. Fig. \ref{FigFormsSIDF}(a) shows that the distance $d(A,B)$ of non-neighboring patches $A$ and $B$ in SIDF can be 
different. Theoretically, the distance can be arbitrary. But it results that the number of all possible SIDF is very large. Because a pedestrian is 
horizontally symmetrical in a loose sense, we restrict the location 
$l(B)$ of $B$ in the interval of the locations $l(A)$ and $l(A')$ where $A'$ is 
the horizontal mirror of $A$. That is, $l(B)\in [l(A),l(A')]$. As 
demonstrated in Fig. \ref{FigLocBSIDF}, $l(B)$ is randomly sampled from $[l(A),l(A')]$ in our experiments. 

Both Figs. \ref{FigFormsSIDF}(b) and (c) show varying sizes of patches. But in Fig. \ref{FigFormsSIDF}(b) both 
two non-neighboring patches equally vary with size (scale) whereas in Fig. 
\ref{FigFormsSIDF}(c) only one patch varies its size. It's good enough for letting $A$ and $B$ have the different width but the same height. Figs. \ref{FigLocBSIDF}(c) and (d) also give an example of the different widths of patches $A$ and $B$. Fig. \ref{FigFormsSIDF}(d) shows SIDF with varying aspect ratio.

\begin{figure}[!t]
\centering
\includegraphics[width=3in]{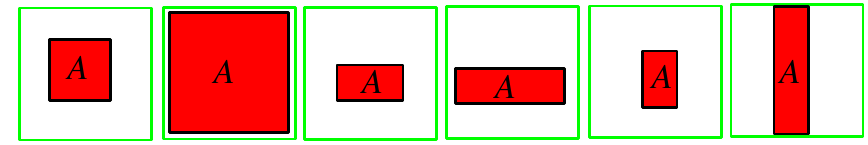}
\caption{The size of patch $A$ is allowed to change inside the maximum square indicated by green squares.}
\label{FigSIDFMax}
\end{figure}

The size of a patch (e.g., patch $A$ in Fig. \ref{FigLocBSIDF}(a)) is allowed to change in a reasonable range. In this paper, the variation of a patch is limited to a maximum square. In other words, the sizes of both patches $A$ and $B$ are allowed to be not larger than that of the maximum square. The green squares in Fig. \ref{FigSIDFMax} are maximum squares and patches have to be inside them. A typical maximum square is of size $8\times 8$ cells (1 cell=$2\times2$ pixels). 

Suppose that the side-inner difference feature $f(A,B)$ consists of two patches $A$ and $B$ (see Fig. \ref{FigLocBSIDF}(a)). The number of pixels of $A$ and $B$ are denoted by $N_A$ and $N_B$. Let $S_A$ and $S_B$ be the pixel sum in $A$ and $B$, respectively. Then the side-inner difference feature $f(A,B)$ can be calculated by
\begin{equation}
\label{eqSIDF}
f(A,B)=\frac{S_A}{N_A}-\frac{S_B}{N_B},
\end{equation}
where $N_A$ and $N_B$ are used for normalization. 

\subsection{Symmetrical similarity features inspired by shape symmetry}
As stated in Section 3.1, the shape of pedestrian is symmetrical. Thus, 
patches $A$ and $A'$ in Fig. \ref{FigLocBSIDF} have the 
similar characteristic. The symmetrical similarity features 
$f(A,A')$ of patches $A$ and $A'$ can be calculated by the following equation:
\begin{equation}
\label{eqSSFori}
f(A,A')=|f_A-f_{A'}|,
\end{equation}
where $f_A$ and $f_{A'}$ represent the features of patches 
$A$ and $A'$ (e.g., histogram features and local mean features). For the 
computational efficiency, we just use the local mean features to represent the patches. 
Namely, $f_A=S_A/N_A$ and $f_{A'}=S_{A'}/N_{A'}$. Then, Eq. (\ref{eqSSFori}) can 
be written as the following:
\begin{equation}
\label{eqSSFori1}
f(A,A')=|\frac{S_A}{N_A}-\frac{S_{A'}}{N_{A'}}|.
\end{equation}

However, due to the changes of the pedestrian posture, the pedestrian symmetry  is relatively loose. It results that Eq. (\ref{eqSSFori1}) is very sensitive to 
the pedestrian deformation. 

\begin{figure}[!t]
\centering
\subfloat[]{\includegraphics[width=0.6in]{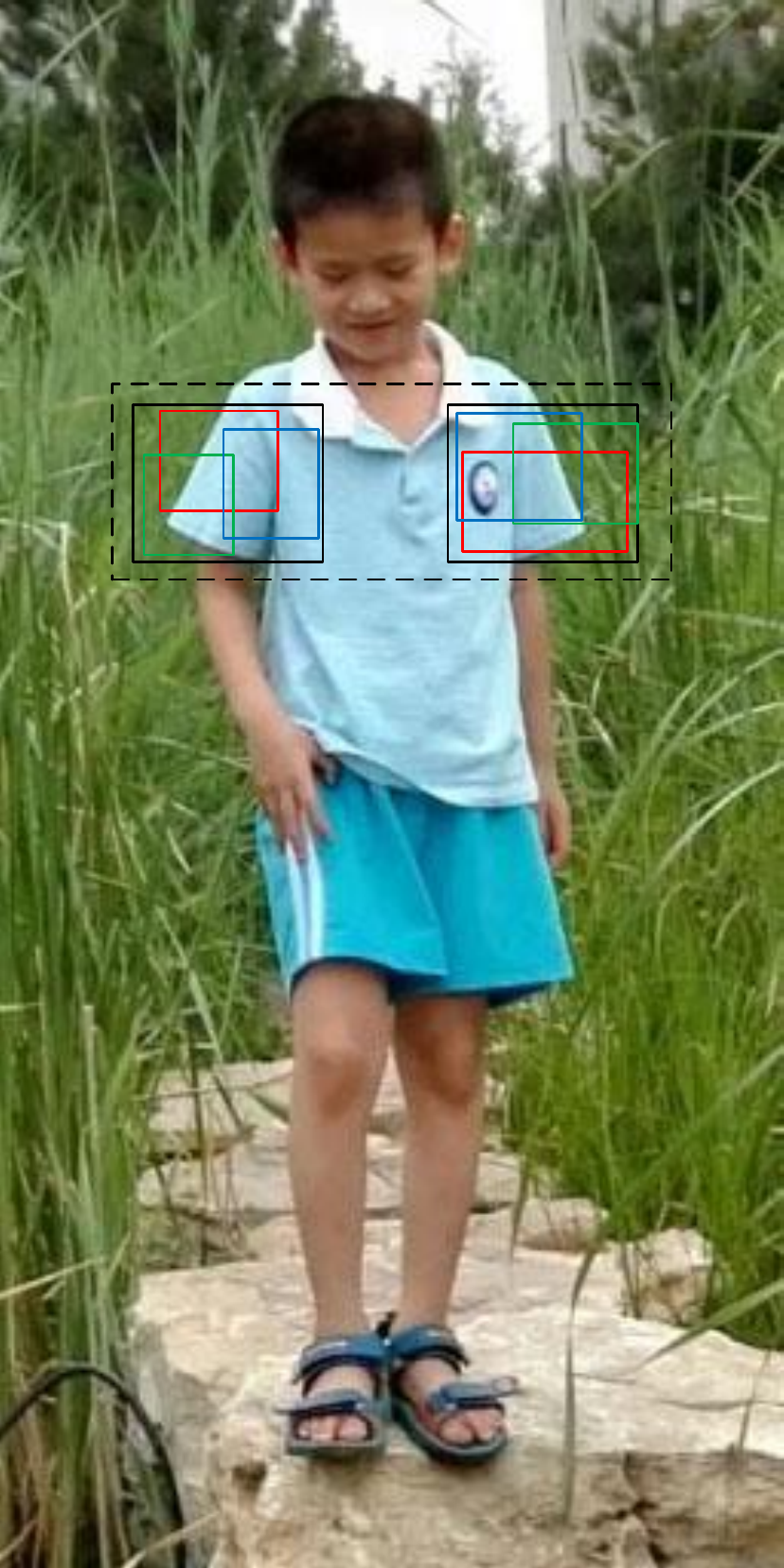}
\label{FigSSF(a)}}
\hfil
\subfloat[]{\includegraphics[width=1.6in]{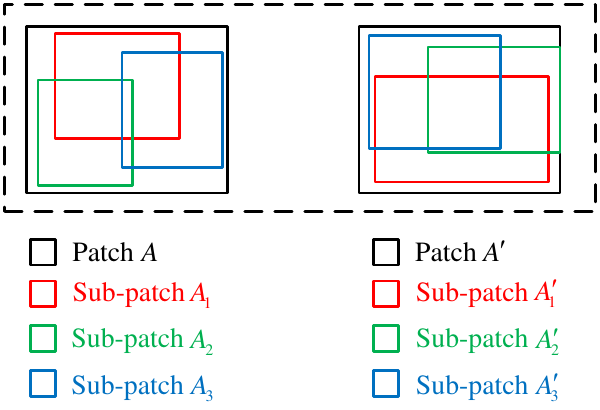}
\label{FigSSF(b)}}
\caption{Examples of symmetrical similarity features. (a) is an example of SSF located in the pedestrian. (b) shows a specific form of SSF.}
\label{FigSSF}
\end{figure}

To eliminate the above influence caused by pedestrian deformation, we 
replace the mean features of patches by the max-pooling features \cite{Wang_Regionlets_ICCV_2013}. In 
Fig. \ref{FigSSF}, two symmetrical patches $A$ and $A'$ are 
represented by three different color sub-patches, respectively. For 
examples, patch $A$ consists of three sub-patches $A_1$, $A_2$, and $A_3$. The sub-subpatches are randomly generated inside the patch $A$. The size and aspect ratio of them can arbitrary, whereas the 
area of them should be larger than half of patch $A$. Then, the feature value of patch 
$A$ is set as the maximum of mean values of three sub-patches. It can be 
expressed as:
\begin{equation}
\label{eqMaxPooling}
f_M(A)=\max_{i=1,2,3}{\frac{S_i}{N_i}}.
\end{equation}
Note that the maximum is replaced by minimum in L and V channel images. Then, the 
symmetrical symmetry features $f(A,A')$ of patches $A$ and $A'$ is calculated by the following equation:
\begin{equation}
\label{eqSSFPooling}
f(A,A')=|f_M(A-f_M(A'))|.
\end{equation}

The size of the symmetrical patches $A$ and $A'$
is allowed to change in a reasonable range, which varies from $6\times 6$ 
cells to 12$\times $12 cells. As the symmetry in pedestrians mainly 
exists in L, U, V, and G channel images, we only use the above channel images to generate SSF.

\subsection{Neighboring features}
\begin{figure}[!t]
\centering
\subfloat[Local mean features.]{\includegraphics[width=3in]{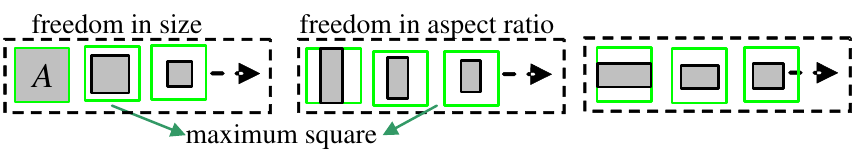}
\label{FigNF(a)}}
\vfil
\subfloat[Neighboring difference features.]{\includegraphics[width=3in]{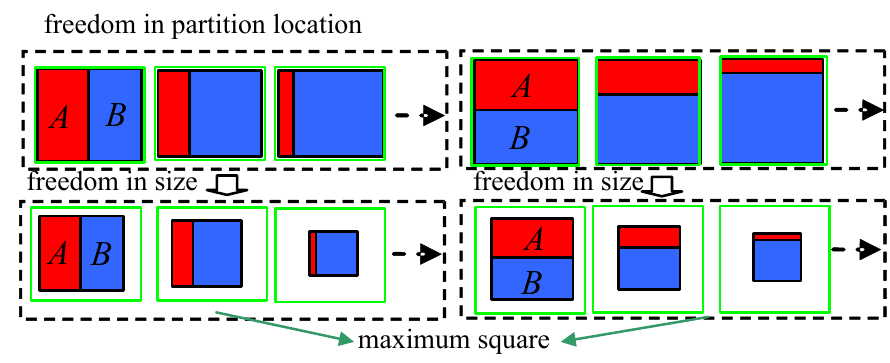}
\label{FigNF(b)}}
\caption{Some possible forms of neighboring features. The green squares are called maximum squares.}
\label{FigNF}
\end{figure}

In fact, both non-neighboring and neighboring features are crucial for pedestrian detection. In this section, we propose to form the pool of neighboring features by using local mean features (see Fig. \ref{FigNF}(a)) and neighboring difference features (see 
Fig. \ref{FigNF}(b)) with enough freedom in size, aspect ratio, patch direction, and 
partition location. The left portion of Fig. \ref{FigNF}(a) shows that the size of a 
feature is allowed to vary in a large extent. Patch direction is either 
vertical or horizontal. The patch direction in the middle of Fig. \ref{FigNF}(a) and 
the left portion of Fig. \ref{FigNF}(b) is vertical whereas the direction in the right 
portion of Fig. \ref{FigNF}(a) and the right portion of Fig. \ref{FigNF}(b) is horizontal.

Partition location is illustrated in Fig. \ref{FigNF}(b) which is defined as the 
location where two neighboring patches intersect. Assigning freedom in 
partition location strengthens representative and discriminative ability of 
the features. 

To avoid the large number of features, we specify a maximum square. 
The sizes of local mean features and neighboring difference features are 
allowed to be not larger than the size of the maximum square. The green squares in 
Fig. \ref{FigNF} are maximum squares. As stated in Section 3.2, a typical size of 
the maximum square is 8$\times $8 cells. 

The neighboring features illustrated in Fig. \ref{FigNF} are suitable to be computed 
with integral image. Hence the feature extraction process is very efficient. 
Note that neighboring difference features can be calculated using the same 
formula (i.e., Eq. (\ref{eqSIDF})) of non-neighboring features.

In our method, both the neighboring (i.e., local mean features and 
neighboring difference features) and non-neighboring features (i.e., SIDF and SSF) are used as input of decision forests and AdaBoost.

\section{Experiments}
The public Caltech pedestrian dataset \cite{Caltech,Dollar_PD_PAMI_2012} and 
INRIA dataset \cite{Dalal_HOG_CVPR_2005} are employed for evaluation. In 
the INRIA dataset, there are 1237 pedestrian images used for training and 
288 pedestrian images used for evaluation.

The Caltech pedestrian dataset is more challenging than the INRIA dataset 
and hence has become a benchmark. It consists of approximately 10 hours of 
$640\times 480$ 30Hz video \cite{Caltech}. The 10 hours data consists of 11 
videos with the first 6 videos are used for training and the last 5 videos 
for testing. The standard 
positive training data is formed by sampling one image out of each 30 
sequential frames. To enlarge the number of training samples, we sample a frame from every two 
or ten frames instead of every 30 frames. The resulting 
training sets are called Caltech 2x and Caltech 10x \cite{Zhang_FCF_CVPR_2015}. Whenever Caltech 2x training set or Caltech 10x training set is used, the testing dataset is the same. The testing dataset consists of 4024 frames among which there are 1014 positive images.

\subsection{Self-comparison using the Caltech 2x training data }

Before comparing with the state-of-the-art methods, experimental results on 
Caltech 2x dataset are reported to show how the proposed method works and 
the importance of each component of the proposed method. Note that the 
Caltech 2x training set instead of Caltech10x training set is used. 

The experimental setup is as follows. Classical 10 channel images (i.e., 
HOG+LUV) are used for generating features. The final classifier consists of 
4096 level-2 decision trees. The classifier is learned by five rounds, where 
the numbers of trees in subsequent rounds are 32, 128, 512, 2048, and 4096, 
respectively. Each tree is built by randomly sampling 1/32 of features from 
the large pool of features. 5000 hard negatives are added after each round 
and the cumulative negatives are limited to 15000. The stride of sliding 
windows is 4 pixels. The model size is 64$\times $128, which consists of 
2048 cells (1 cell=2x2 pixels). As the pedestrian is generally taller than 50 
pixels, each testing image is upsampled by one octave. 

\begin{table}
\begin{center}
\centering
\begin{tabular}{|c|c|c|c|c|c|}
\hline
Channel & L & U & V & G & 6 Oriented gradients \\
\hline
Method & $\frac{x-\mu}{\sigma}$ & \multicolumn{2}{c|}{$x$}   & \multicolumn{2}{c|}{$\frac{x}{\mu_G}$}\\
\hline
\end{tabular}
\end{center}
\caption{Channel-Specific Normalization}
\end{table}

In NNNF (a.k.a., NNF+NF), both Non-Neighboring Features (NNF) and 
Neighboring Features (NF) are employed. In the NNF, there are two types of 
non-neighboring features: SIDF and SSF. NF+SIDF or NF+SSF mean that 
the neighboring features are combined with only one type of non-neighboring features (i.e., 
SIDF or SSF). In SIDF and NF, the channel-specific 
normalization can be used in Table 1. In Table 1, $x$ is a feature in a detection window. $\mu$ and $\sigma$ are the mean and variance of the features in the detection window. $\mu_G$ is the mean of G channel. Because U and V channels are relatively stable to variations in illumination, we do not perform normalization. We denote NNNF-No the 
method which is the same as NNNF except that no normalization is conducted 
in SIDF and NF.

\begin{figure}[!t]
\centering
\includegraphics[width=2.2in]{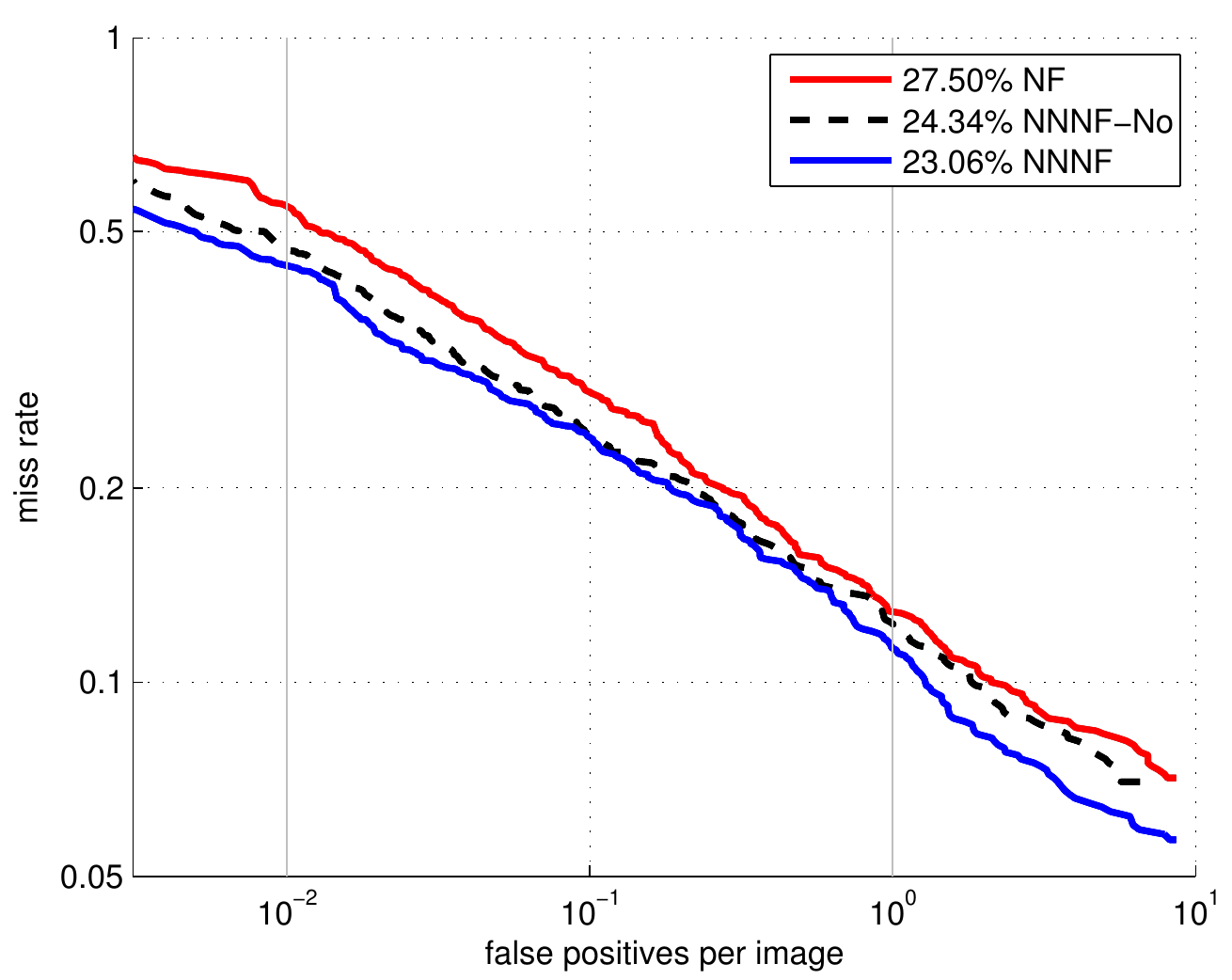}
\caption{ Self-comparison: ROC curves of NF, and NNNF-No, and NNNF on the Caltech dataset.}
\label{FigComOurs}
\end{figure}

The ROC curves of NF, NNNF-No and NNNF are shown in Fig. \ref{FigComOurs}. It is seen that 
the performance of NNNF is systematically better than that of NF, meaning 
that incorporating NNF is useful for improving detection performance. 
Meanwhile, one can observe that NNNF-No is inferior to NNNF. NNNF employs 
channel-specific normalization in NF and SIDF whereas 
NNNF-No does not perform normalization. So it is concluded that pedestrian 
detection benefits from the proposed channel-specific normalization. 

\begin{table}
\begin{center}
\begin{tabular}{c c c }
\hline
Method & MR        &$\Delta$ MR \\
\hline
NF     & 27.50\%   & N/A \\
NF+SIDF & 25.67\%  & +1.83\% \\
NF+SSF & 25.20\%   & +2.30\% \\
NNNF-No & 24.34\%   & +3.16\% \\
NNNF   & 23.06\%      & +4.44\%\\
\hline
\end{tabular}
\end{center}
\caption{Comparison of Log-average Miss Rates}
\end{table}

The above observation can also be seen from Table 2 where the log-average 
miss rates are given. The miss rates of NNNF 
(i.e., NNF+NF), NNNF-No, and NF are 23.06{\%}, 24.34{\%}, and 27.50{\%}, 
respectively. The miss rate of NNF+NF is 4.44{\%} smaller than 
that of NF. So it is said that non-neighboring features contribute 
significantly for improving detection performance. Specifically, NF+SIDF and 
NF+SSF outperform NF by 1.83{\%} and 2.30{\%}, respectively. NNNF 
outperforms NNNF-No by 1.28{\%}. Though the contribution of channel-specific 
normalization is not as significant as non-neighboring features, it is 
steadily helpful for improving detection performance.

\begin{figure}[!t]
\centering
\includegraphics[width=2.5in]{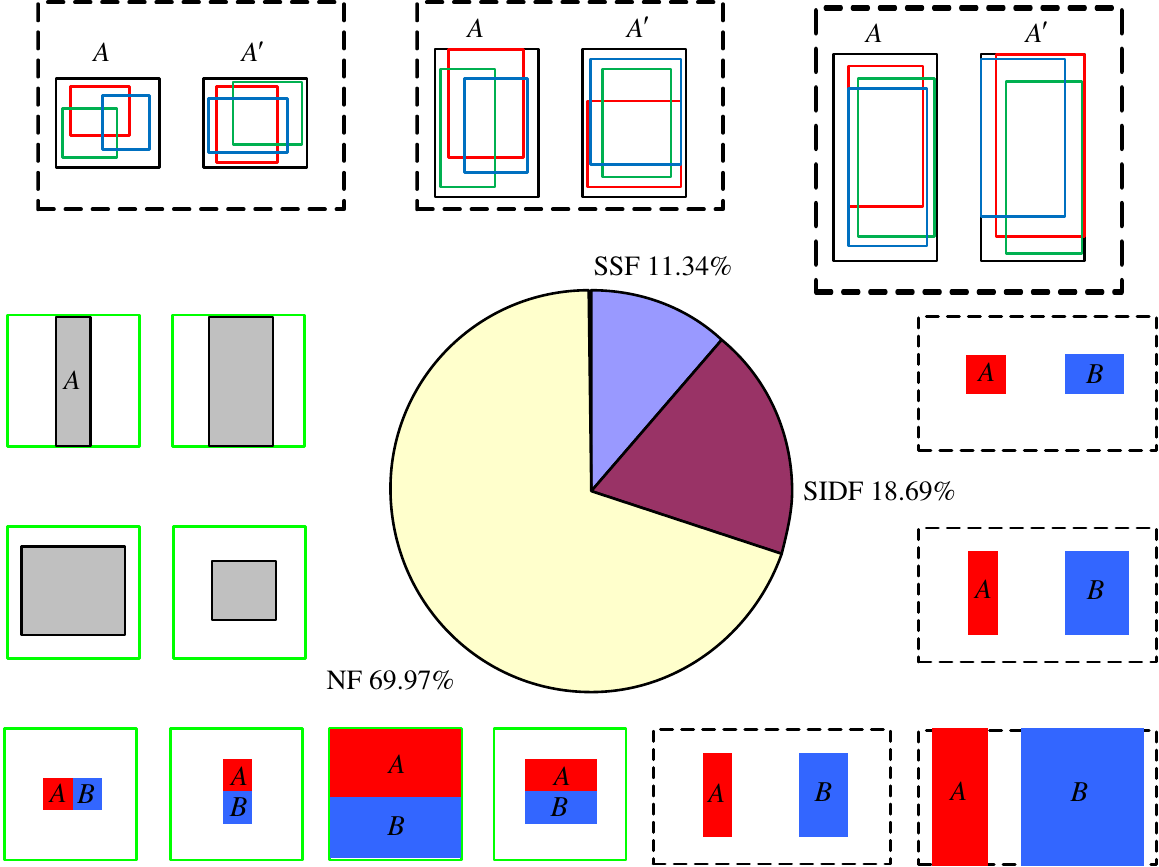}
\caption{ Among all the selected features, about 30\% are non-neighboring features and 70\% are neighboring features. Some representative non-neighboring and neighboring features also shown.}
\label{FigPerNNNF}
\end{figure}

Totally, 12288 features are selected, which consist of 3690 
non-neighboring features and 8598 neighboring features. Among 
non-neighboring features, there are 2297 side-inner difference features 
(SIDF) and 1393 symmetrical similarity features (SSF). That is, the 
proportions of SIDF, SSF, and NF are approximately 18.69{\%}, 11.34{\%} and 
69.97{\%} (see Fig. \ref{FigPerNNNF}). We can conclude that non-neighboring features are 
complementary to neighboring features. Several representative forms of 
non-neighboring (SIDF and SSF) and neighboring features (NF) are also shown 
in Figs. \ref{FigPerNNNF}. 

\begin{figure}[!t]
\centering
\subfloat[]{\includegraphics[width=0.7in]{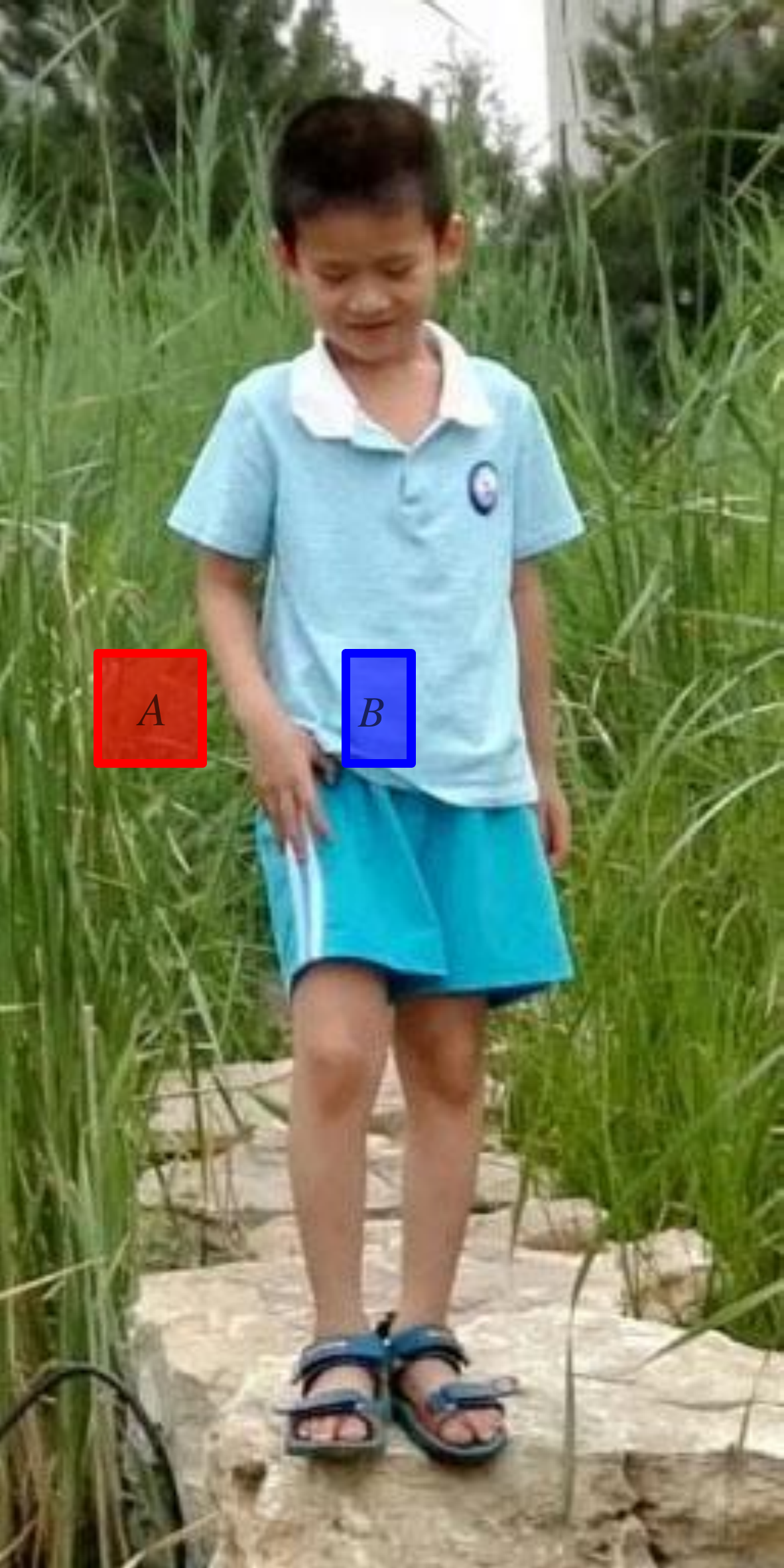}
\label{FigSIDFSSF(a)}}
\hfil
\subfloat[]{\includegraphics[width=0.7in]{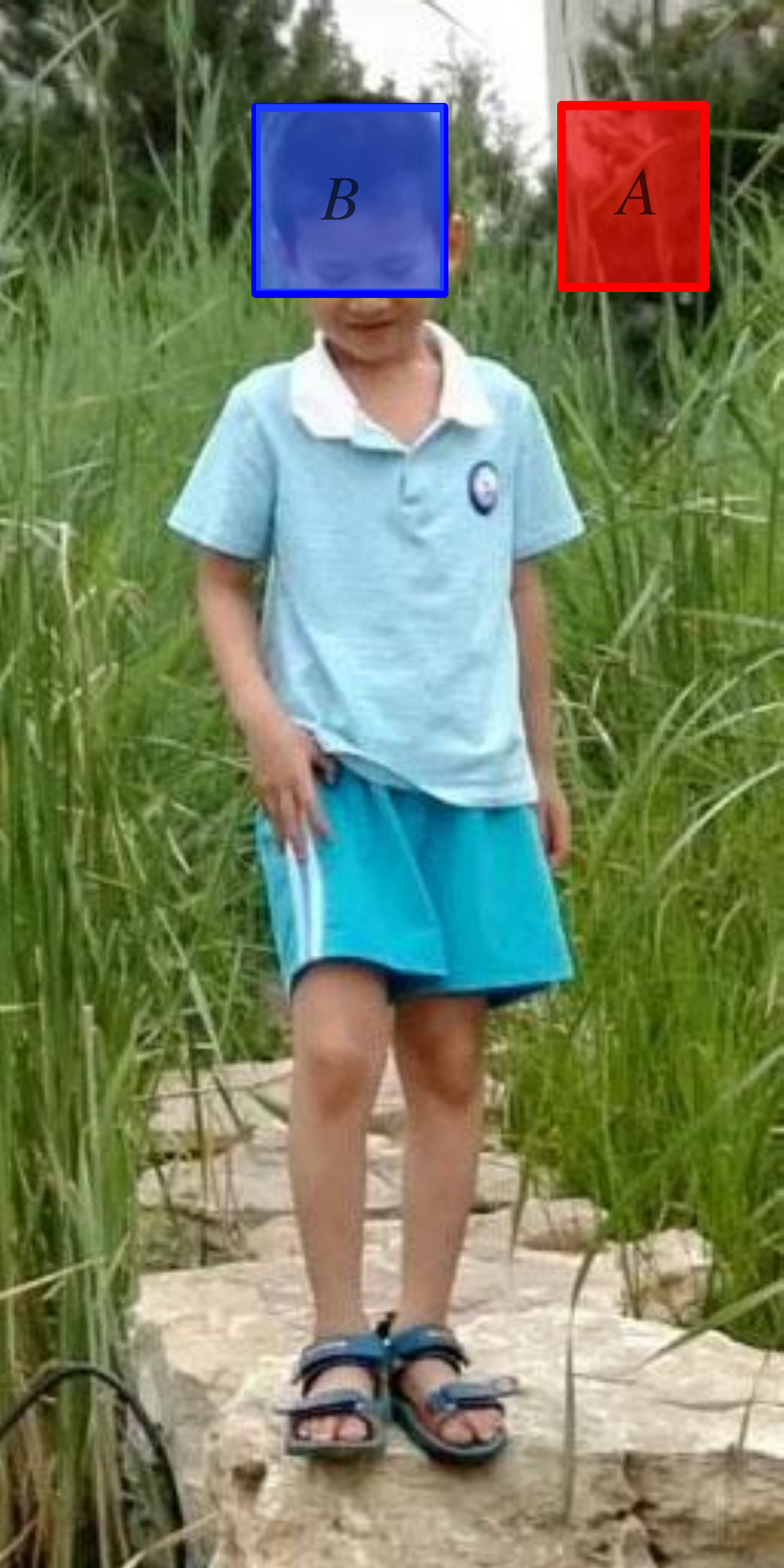}
\label{FigSIDFSSF(b)}}
\hfil
\subfloat[]{\includegraphics[width=0.7in]{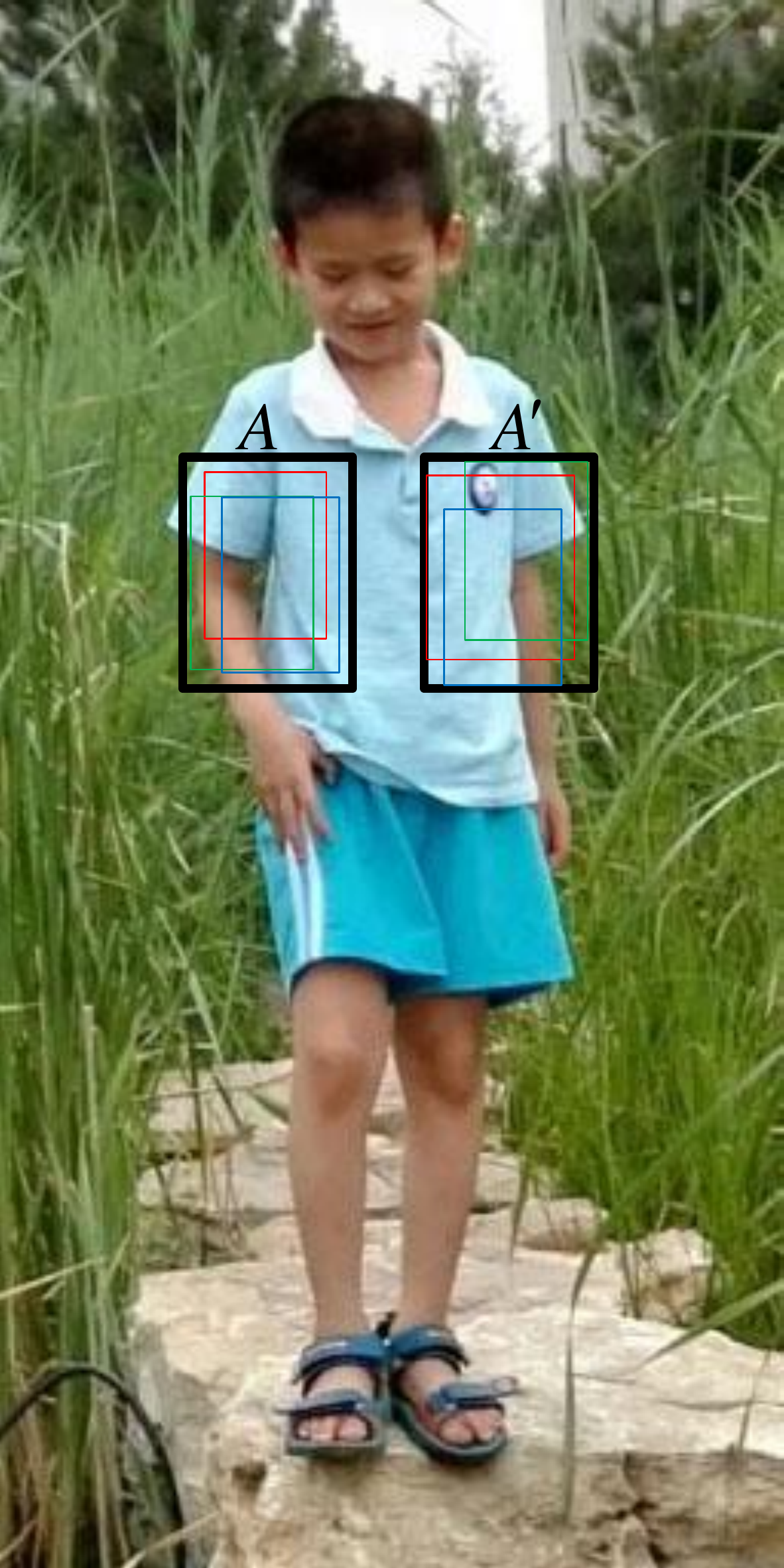}
\label{FigSIDFSSF(c)}}
\hfil
\subfloat[]{\includegraphics[width=0.7in]{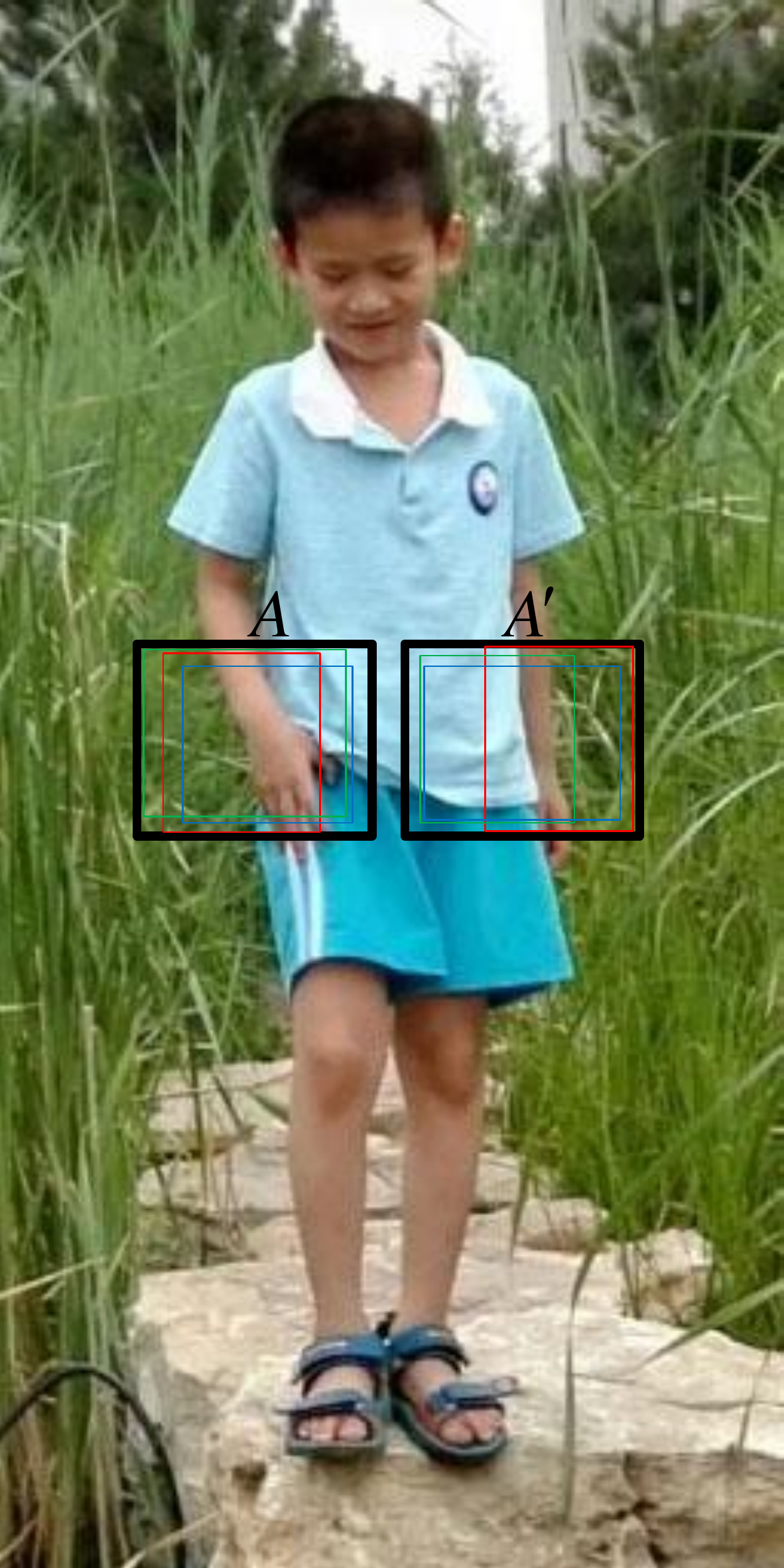}
\label{FigSIDFSSF(d)}}
\caption{Several selected non-neighboring features. The first two features are SIDF, and the last two features are SSF.}
\label{FigSIDFSSF}
\end{figure}

In Fig. \ref{FigSIDFSSF}, the representative non-neighboring features are also visualized 
on pedestrian images. The first two images show the side-inner difference 
features, and the last two images show the symmetrical similarity features.

\begin{figure}[!t]
\centering
\subfloat[]{\includegraphics[width=1.0in]{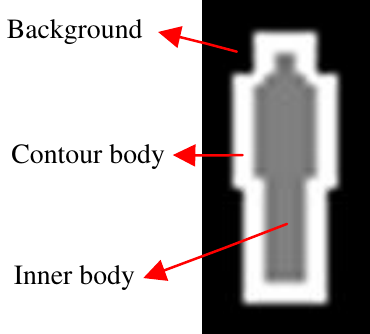}
\label{FigPerSIDF(a)}}
\hfil
\subfloat[]{\includegraphics[width=1.6in]{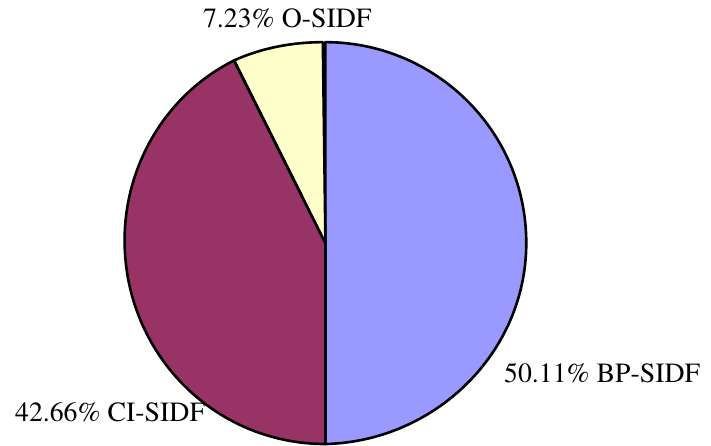}
\label{FigPerSIDF(b)}}
\caption{CI-SIDF, BP-SIDF, and O-SIDF features. (a) The ternary model of pedestrians. (b) The portions of SIDF.}
\label{FigPerSIDF}
\end{figure}

%

In fact, SIDF features can be categorized into the following three types: 
1) A SIDF feature is called Contour-Inner SIDF (CI-SIDF) feature if one of 
its patch is located on the pedestrian contour and the other is 
located inside the pedestrian; 2) A SIDF feature is called Background-Pedestrian 
SIDF (BP-SIDF) feature if one of its patch is on the background and the 
other patch is inside or on the contour of a pedestrian; and 3) A SIDF 
feature different from CI-SIDF and BP-SIDF features is called Other SIDF 
(O-SIDF) feature. To know the proportions of the three types of SIDF 
features, a ternary model (Fig. \ref{FigPerSIDF}(a)), consisting of background, contour body, 
and inner body, is created according to average appearance (e.g., Fig. \ref{FigAvgPed}(d)) 
of pedestrians. All the 2297 selected SIDF features are classified to 
CI-SIDF, BP-SIDF, and O-SIDF by computing the intersection of a SIDF feature 
and the ternary model. The results given in Fig. \ref{FigPerSIDF}(b) indicate that the 
proportions of CI-SIDF, BP-SIDF, and O-SIDF are 42.66{\%}, 50.11{\%}, and 
7.23{\%}, respectively. Fig. \ref{FigPerSIDF}(b) tells that SIDF features not only capture 
the difference the contour of a pedestrian and its inner part but also 
utilize the difference between the background and a pedestrian. Background 
can be regarded as context of a pedestrian image and hence context has been 
proved to be effective in object detection and recognition. It is difficult 
for neighboring features to utilize the context information. 

\subsection{Comparison with state-of-the-art methods on Caltech dataset}
The proposed NNNF method can adopt different levels (depths) decision trees. 
In this section, NNNF-L2 stands for the NNNF method where level-2 trees are 
utilized. The Caltech 2x training data is used for NNNF-L2. All parameters 
in NNNF-L2 are the same as those in Section 4.1. In NNNF-L4, level-4 trees 
are employed. The Caltech 10x training data is used for NNNF-L4. The 
resulting classifier is composed of 4096 level-4 decision trees and each 
tree is built by randomly sampling 1/2 of features from the feature pool. 
The decision trees are obtained after five rounds. In each round, 20000 hard 
negatives are added and the cumulative negatives are limited to 50000. Other 
parameters are the same as those in Section 4.1. 

\begin{figure}[!t]
\centering
\includegraphics[width=2.5in]{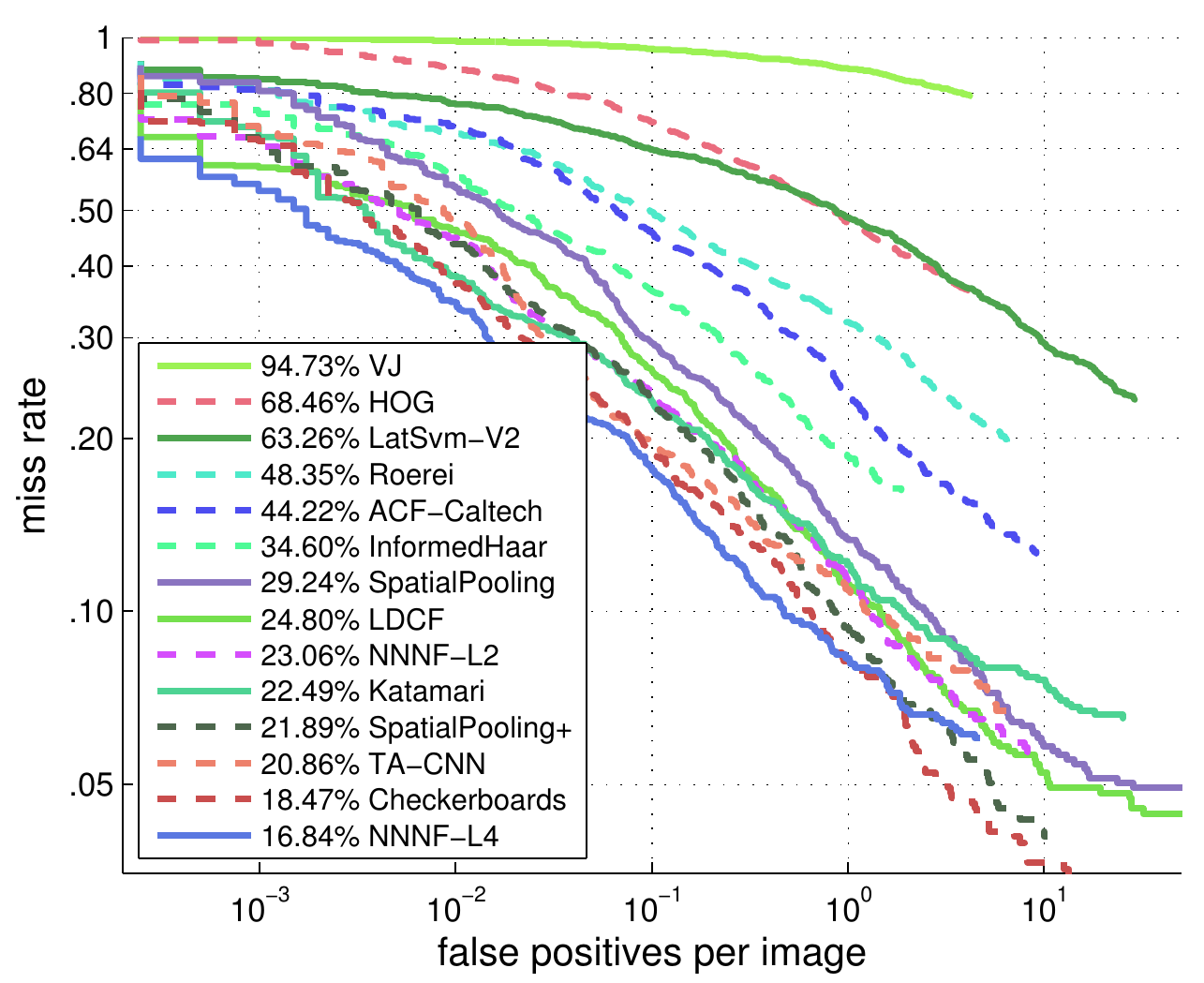}
\caption{Comparison with state-of-the-art methods on the Caltech dataset.}
\label{FigCaltech}
\end{figure}

Fig. \ref{FigCaltech} compares NNNF-L2 and NNNF-L4 with the state-of-the-art 
methods. The 
curves of ACF-Caltech \cite{Dollar_PD_PAMI_2012} are obtained when they are trained on the Caltech 
training set. The models of VJ \cite{Viola_RoFace_IJCV_2004}, HOG \cite{Dalal_HOG_CVPR_2005}, LatSvm-V2 \cite{Felzenszwalb_DPM_CVPR_2008}, and Roerei \cite{Benenson_SquareChns_CVPR_2013} are trained on the INRIA dataset. The 
curves of other methods are obtained when the training set is Caltech 10x. 
They all utilize the Caltech testing set for evaluation. 

The following observations can be seen from Fig. \ref{FigCaltech}. Even the small Caltech 
2x training dataset is used, the proposed NNNF-L2 is better than LDCF \cite{Nam_LDCF_NIPS_2014} whose 
models are trained from the large Caltech10x dataset. Specifically, the 
log-average miss rate of NNNF-2 is 23.06{\%}. 

It can also be seen from Fig. \ref{FigCaltech} that the proposed NNNF-L4 is superior to 
all other methods. The log-average 
miss rate of NNNF-L4 is as small as 16.84{\%} whereas the log-average miss 
rate of TA-CNN \cite{Tian_Ta_CVPR_2015} and Checkboards \cite{Zhang_FCF_CVPR_2015} are 20.86{\%} and 18.47{\%}, 
respectively. Though the proposed non-neighboring and neighboring features 
are much simpler than those in CNN and Checkboards, they result in better detection results. 

\begin{figure}[!t]
\centering
\includegraphics[width=2.3in]{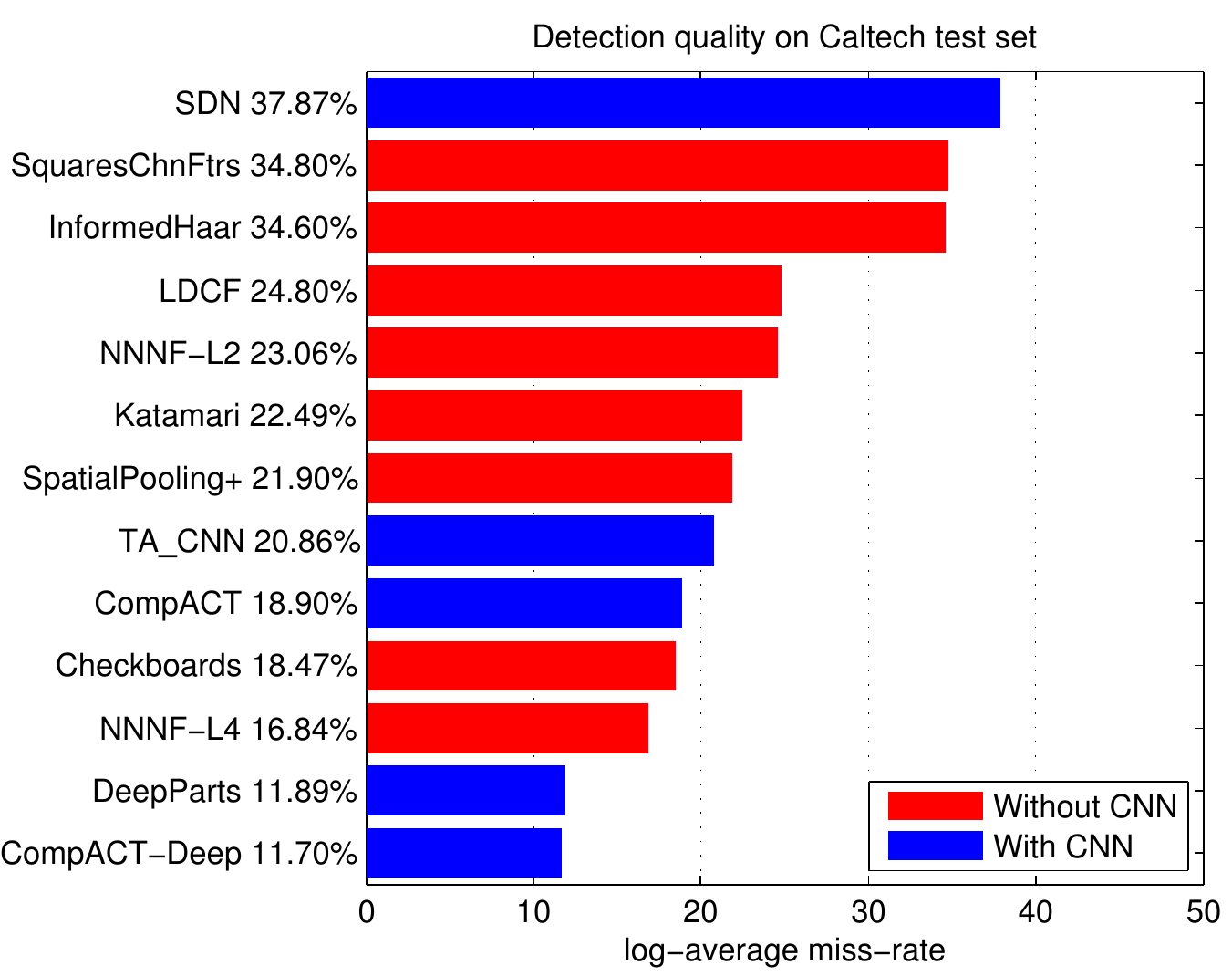}
\caption{Miss rate of the state-of-the-art methods. The methods with blue bars are based on CNN. The methods with red bars are not using CNN.}
\label{FigHatch}
\end{figure}

According to whether using CNN or not, Fig. \ref{FigHatch} divides the state-of-the-art 
methods into two classes. The methods with red bars do not use CNN. NNNF-L4 achives the best detection performance, outperforming Checkboards \cite{Zhang_FCF_CVPR_2015} by 1.63\%. The methods with blue bars are based on CNN. CompACT-Deep \cite{Cai_DeepPed_ICCV_2015} achieves the lowest miss rate (i.e., 11.70\%) by combination of some local channel features (e.g., ACF \cite{Dollar_PD_PAMI_2012}, Checkboards \cite{Zhang_FCF_CVPR_2015}, and LDCF \cite{Nam_LDCF_NIPS_2014}) and deep features (e.g., VGG \cite{Simonyan_VGG_arxiv_2014}). Though CompACT-Deep \cite{Cai_DeepPed_ICCV_2015} has a better performance than NNNF-L4, the improvement of CompACT-Deep are based on very deep CNN model (i.e., VGG). When only using the above local features and small CNN, CompACT can only achieve 18.9\%, which is inferior to NNNF-L4. It means that NNNF-L4 are much more effective than the local features used in CompACT. Moveover, Our non-neighboring features are complementary to the features of CompACT. So the non-neighboring features can be combined with CompACT to boost the performance of pedestrian detection.

\begin{figure}[!t]
\centering
\includegraphics[width=3in]{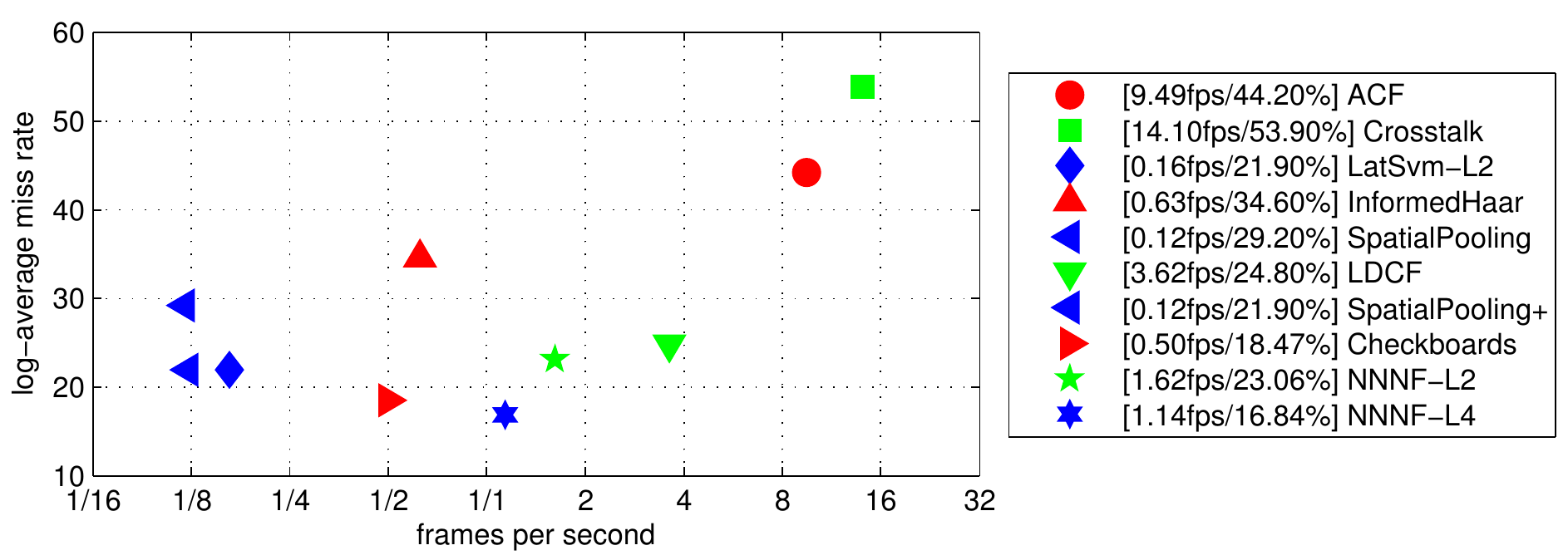}
\caption{Log-average miss rate (MR) versus frames per second (FPS) on the Caltech.}
\label{FigRunTimeCaltech}
\end{figure}

The log-average miss rates and frames per second of the methods without CNN are visualized in Fig. \ref{FigRunTimeCaltech}. It is 
desirable if miss rate is as small as possible and FPS is as large as 
possible. So Fig. \ref{FigRunTimeCaltech} implies that the proposed NNNF-L4 achieves the best 
tradeoff between miss rate and FPS. The log-average rate of NNNF-L4 is superior to 
that of Checkboards, and it is also 2.28 times faster than Checkboards. Note that the detection speed is 
measured on a computer with an Intel i7 CPU and a 640$\times $480 image 
with the height of a pedestrian not less than 50 pixels. GPU is not used. 

\subsection{Comparison with state-of-the-art methods on the INRIA dataset}
Experiments are also conducted on the INRIA dataset. Because pedestrian 
height in both the training and testing sets are larger than 100 pixels, we 
train a model with 64$\times $128 pixels. The model consists of 2048 level-3 
decision trees. Other parameters are the same as those in Section 4.1. 

\begin{figure}[!t]
\centering
\includegraphics[width=2.5in]{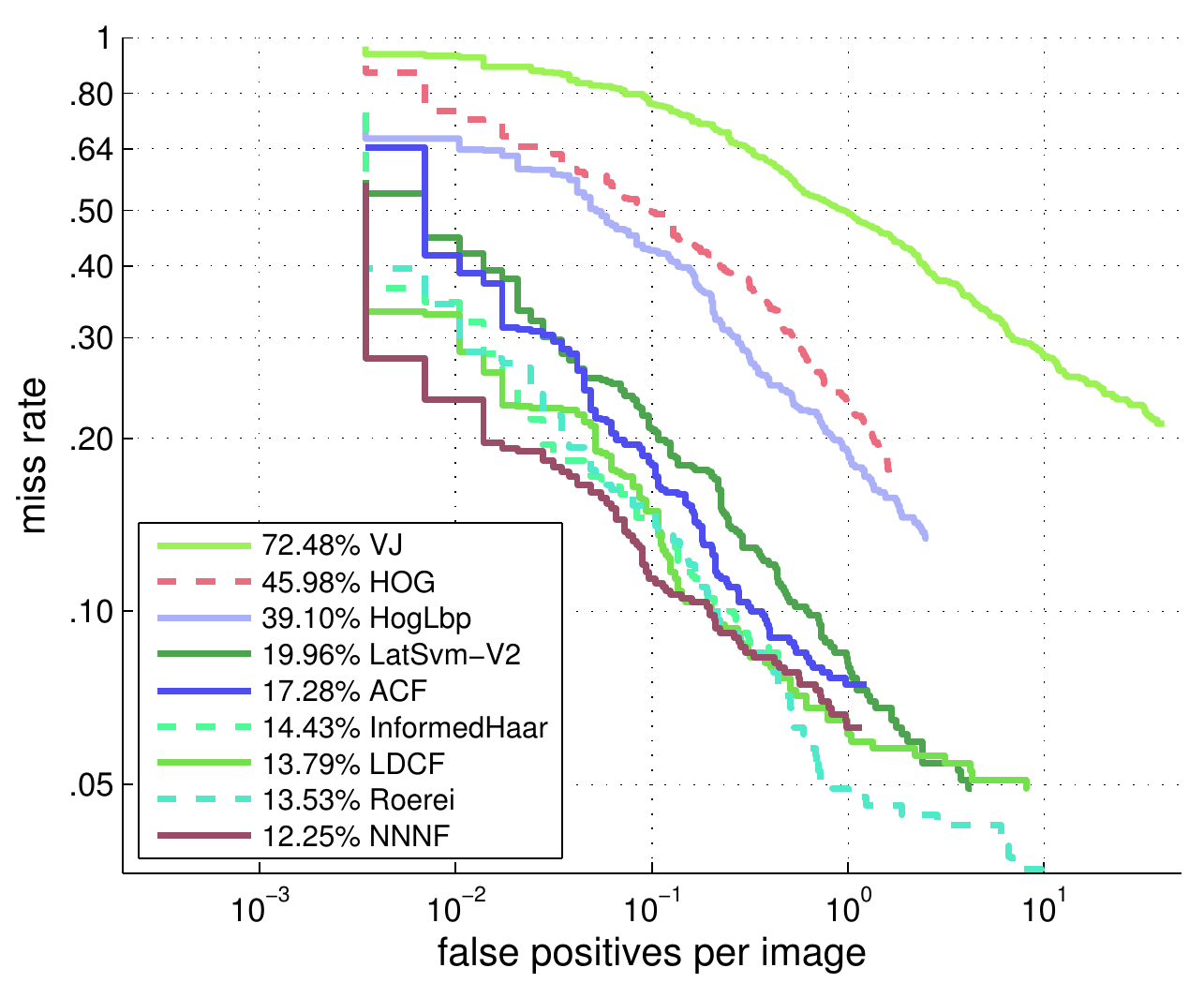}
\caption{Comparison with state-of-the-art methods on the INRIA dataset.}
\label{FigINRIA}
\end{figure}

Experimental results are shown in Fig. \ref{FigINRIA}. It can be observed that NNNF achieves the best performance (log-average miss rate is 
12.25{\%}). The miss rate of NNNF is 7.71{\%}, 5.03{\%}, and 2.18{\%}
lower than that of LatSvm-V2 \cite{Felzenszwalb_DPM_CVPR_2008}, ACF \cite{Dollar_ACF_PAMI_2014}, and InformedHaar \cite{Zhang_Info.Haar_CVPR_2014}. NNNF outperforms LDCF 
\cite{Nam_LDCF_NIPS_2014} by 1.54{\%}. The advantage 
of NNNF over LDCF \cite{Nam_LDCF_NIPS_2014} and Roerei \cite{Benenson_SquareChns_CVPR_2013} is more remarkable for the complex Caltech 
dataset than for the simple INRIA dataset. 

\begin{figure}[!t]
\centering
\includegraphics[width=3in]{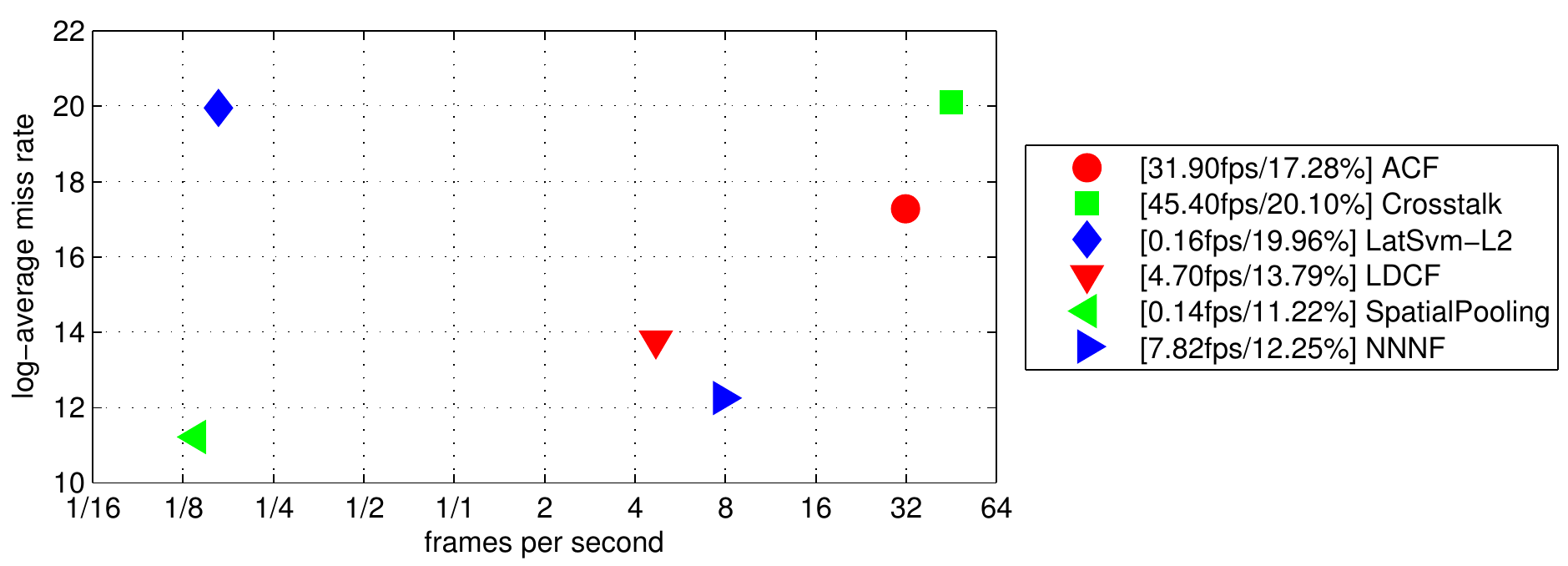}
\caption{Log-average miss rate (MR) versus frames per second (FPS) on the INRIA.}
\label{FigRunTimeINRIA}
\end{figure}

The comparison of detection speed and miss rate of different methods is 
given in Fig. \ref{FigRunTimeINRIA}. The image to be detected has 640$\times $480 pixels and the 
height of a pedestrian is not less than 100 pixels. One can see that NNNF outperforms in terms of log-average miss rate all the 
methods except Spatialpool. Though slightly lower than the miss rate of 
Spatialpool, NNNF is 55.86 times faster than SpatialPool \cite{Paisitkriangkrai_SpatialPool_ECCV_2014}. Therefore, our 
method is able to get the best tradeoff between miss rate and detection 
speed.

\section{Conclusion}
In this paper, we have presented an effective and efficient pedestrian detection method. The main contribution lies in the proposed two types of non-neighboring features (NNF): side-inner difference features (SIDF) and symmetrical similarity features (SSF) which were found to be complementary to the proposed neighboring features (NF). SIDF features characterize not only the difference between contour of a pedestrian and its inner part but also the difference of the background and pedestrian. SSF can capture the symmetrical similarity of pedestrian shape. Though the forms of the proposed NNF and NF features are very simple, combining them results in the best tradeoff between miss rate and frames per second.

It is noted that the proposed non-neighboring features (i.e., NNF), the neighboring features (e.g., Checkboards), and CNN features (e.g., VGG) are complementary. In the future work, we will focus on combing NNF with them and investigating whether or not this combination is able to improve the performance of pedestrian detection.

{\small
\bibliographystyle{ieee}
\bibliography{egbib}

\begin{thebibliography}{10}\itemsep=-1pt

\bibitem{Caltech}
\url{http://www.vision.caltech.edu/Image\_Datasets/CaltechPedestrians/}.

\bibitem{Appel_Pruning_ICML_2013}
R.~Appel, T.~Fuchs, P.~Doll{\'a}r, and P.~Perona.
\newblock Boosting decision trees-pruning underachieving features early.
\newblock {\em Proc. Int¡¯l Conf. Machine Learning}, 2013.

\bibitem{Benenson_SquareChns_CVPR_2013}
R.~Benenson, M.~Mathias, T.~Tuytelaars, and L.~V. Gool.
\newblock Seeking the strongest rigid detector.
\newblock {\em Proc. IEEE Conf. Computer Vision and Pattern Recognition}, 2013.

\bibitem{Benenson_TenYears_ECCV_2014}
R.~Benenson, M.~Omran, J.~Hosang, and B.~Schiele.
\newblock Ten years of pedestrian detection, what have we learned?
\newblock {\em Proc. European Conf. Computer Vision}, 2014.

\bibitem{Bourdev_SoftCascade_CVPR_2005}
L.~Bourdev and J.~Brandt.
\newblock Robust object detection via soft cascade.
\newblock {\em Proc. IEEE Conf. Computer Vision and Pattern Recognition}, 2005.

\bibitem{Cai_DeepPed_ICCV_2015}
Z.~Cai, M.~Saberian, and N.~Vasconcelos.
\newblock Learning complexity-aware cascades for deep pedestrian detection.
\newblock {\em Proc. IEEE Int'l Conf. Computer Vision}, 2015.

\bibitem{Dalal_HOG_CVPR_2005}
N.~Dalal and B.~Triggs.
\newblock Histograms of oriented gradients for human detection.
\newblock {\em Proc. European Conf. Computer Vision}, 2005.

\bibitem{Dollar_ACF_PAMI_2014}
P.~Doll{\'a}r, R.~Appel, S.~Belongie, and P.~Perona.
\newblock Fastest feature pyramids for object detection.
\newblock {\em IEEE Trans. on Pattern Analysis and Machine Intelligence},
  36(8):1532--1545, 2014.

\bibitem{Dollar_FastestWest_BMVC_2010}
P.~Doll{\'a}r, S.~Belongie, and P.~Perona.
\newblock The fastest pedestrian detector in the west.
\newblock {\em Proc. British Machine Vision Conference}, 2010.

\bibitem{Dollar_ICF_BMVC_2009}
P.~Doll{\'a}r, Z.~Tu, P.~Perona, and S.~Belongie.
\newblock Integral channel features.
\newblock {\em Proc. British Machine Vision Conference}, 2009.

\bibitem{Dollar_PD_PAMI_2012}
P.~Doll{\'a}r, C.~Wojek, B.~Schiele, and P.~Perona.
\newblock Pedestrian detection: An evaluation of the state of the art.
\newblock {\em IEEE Trans. Pattern Analysis and Machine Intelligence},
  34(4):743--761, 2012.

\bibitem{Felzenszwalb_CascadeDPM_CVPR_2010}
P.~Felzenszwalb, R.~Girshick, and D.~McAllester.
\newblock Cascade object detection with deformable part models.
\newblock {\em Proc. IEEE Conf. Computer Vision and Pattern Recognition}, 2010.

\bibitem{Felzenszwalb_DPM_CVPR_2008}
P.~F. Felzenszwalb, D.~McAllester, and D.~Ramanan.
\newblock A discriminatively trained, multiscale, deformable part model.
\newblock {\em Proc. IEEE Conf. Computer Vision and Pattern Recognition}, 2008.

\bibitem{Girshick_RCNN_CVPR_2014}
R.~B. Girshick, J.~Donahue, T.~Darrell, and J.~Malik.
\newblock Rich feature hierarchies for accurate object detection and semantic
  segmentation.
\newblock {\em Proc. IEEE Conf. Computer Vision and Pattern Recognition}, 2014.

\bibitem{Hariharan_DDCC_ECCV_2012}
B.~Hariharan, J.~Malik, and D.~Ramanan.
\newblock Discriminative decorrelation for clustering and classification.
\newblock {\em Proc. European Conf. Computer Vision}, 2012.

\bibitem{Hosang_DeepLook_CVPR_2015}
J.~H. Hosang, M.~Omran, R.~Benenson, and B.~Schiele.
\newblock Taking a deeper look at pedestrians.
\newblock {\em Proc. IEEE Conf. Computer Vision and Pattern Recognition}, 2015.

\bibitem{Krizhevsky_AlexNet_NIPS_2012}
A.~Krizhevsky, I.~Sutskever, and G.~E. Hinton.
\newblock Imagenet classification with deep convolutional neural networks.
\newblock {\em Proc. Advances in Neural Information Processing Systems}, 2012.

\bibitem{Luo_SDN_CVPR_2014}
P.~Luo, Y.~Tian, X.~Wang, and X.~Tang.
\newblock Switchable deep network for pedestrian detection.
\newblock {\em Proc. IEEE Conf. Computer Vision and Pattern Recognition}, 2014.

\bibitem{Nam_LDCF_NIPS_2014}
W.~Nam, P.~Doll{\'a}r, and J.~H. Han.
\newblock Local decorrelation for improved pedestrian detection.
\newblock {\em Proc. Advances in Neural Information Processing Systems}, 2014.

\bibitem{Ouyang_SingAid_CVPR_2013}
W.~Ouyang and X.~Wang.
\newblock Single-pedestrian detection aided by multi-pedestrian detection.
\newblock {\em Proc. IEEE Conf. Computer Vision and Pattern Recognition}, 2013.

\bibitem{Paisitkriangkrai_SpatialPool_arXiv_2014}
S.~Paisitkriangkrai, C.~Shen, and A.~van~den Hengel.
\newblock Pedestrian detection with spatially pooled features and structured
  ensemble learning.
\newblock {\em arXiv}, 2014.

\bibitem{Paisitkriangkrai_SpatialPool_ECCV_2014}
S.~Paisitkriangkrai, C.~Shen, and A.~van~den Hengel.
\newblock Strengthening the effectiveness of pedestrian detection.
\newblock {\em Proc. European Conf. Computer Vision}, 2014.

\bibitem{Sermanet_PedUMFL_CVPR_2013}
P.~Sermanet, K.~Kavukcuoglu, S.~Chintala, and Y.~LeCun.
\newblock Pedestrian detection with unsupervised multi-stage feature learning.
\newblock {\em Proc. IEEE Conf. Computer Vision and Pattern Recognition}, 2013.

\bibitem{Simonyan_VGG_arxiv_2014}
K.~Simonyan and A.~Zisserman.
\newblock Very deep convolutional networks for large-scale image recognition.
\newblock {\em arXiv}, 2014.

\bibitem{Tian_DeepParts_ICCV_2015}
Y.~Tian, P.~Luo, X.~Wang, and X.~Tang.
\newblock Deep learning strong parts for pedestrian detection.
\newblock {\em Proc. IEEE Int¡¯l Conf. Computer Vision}, 2015.

\bibitem{Tian_Ta_CVPR_2015}
Y.~Tian, P.~Luo, X.~Wang, and X.~Tang.
\newblock Pedestrian detection aided by deep learning semantic tasks.
\newblock {\em Proc. IEEE Conf. Computer Vision and Pattern Recognition}, 2015.

\bibitem{Viola_RoFace_IJCV_2004}
P.~Viola and M.~Jones.
\newblock Robust real-time face detection.
\newblock {\em Int'l J. Computer Vision}, 2004.

\bibitem{Wang_Regionlets_ICCV_2013}
X.~Wang, M.~Yang, S.~Zhu, and Y.~Lin.
\newblock Regionlets for generic object detection.
\newblock {\em Proc. IEEE Conf. Computer Vision and Pattern Recognition}, 2013.

\bibitem{Yang_CCF_ICCV_2015}
B.~Yang, J.~Yan, Z.~Lei, and S.~Z. Li.
\newblock Convolutional channel features.
\newblock {\em Proc. IEEE Int'l Conf. Computer Vision}, 2015.

\bibitem{Zhang_MIP_NIPS_2008}
C.~Zhang and P.~Viola.
\newblock Multiple-instance pruning for learning efficient cascade detectors.
\newblock {\em Proc. Advances in Neural Information Processing Systems}, 2008.

\bibitem{Zhang_Info.Haar_CVPR_2014}
S.~Zhang, C.~Bauckhage, and A.~B. Cremers.
\newblock Informed haar-like features improve pedestrian detection.
\newblock {\em Proc. IEEE Conf. Computer Vision and Pattern Recognition}, 2014.

\bibitem{Zhang_FCF_CVPR_2015}
S.~Zhang, R.~Benenson, and B.~Schiele.
\newblock Filtered channel features for pedestrian detection.
\newblock {\em Proc. IEEE Conf. Computer Vision Pattern Recognition}, 2015.

\end{thebibliography}
}

\end{document}